\documentclass[aps,pre,twocolumn,nobibnotes,amsmath,amssymb,superscriptaddress,longbibliography]{revtex4-2}
\usepackage[draft]{changes}
\definechangesauthor[name={ksxu}, color=orange]{xu}
\definechangesauthor[name={ksxu}, color=cyan]{xu1}
\usepackage{lipsum}
\usepackage{hyperref}
\usepackage{multirow}
\usepackage{float}
\usepackage{colortbl}
\usepackage{xcolor}
\usepackage{graphicx}
\usepackage{dcolumn}
\usepackage{bm}
\usepackage{booktabs}
\usepackage{makecell}
\usepackage{amsmath}
\usepackage{cases}
\usepackage{float}
\usepackage{lineno}

\begin{document}
	\preprint{APS/123-QED}

\title{Impact of leaky dynamics on predictive path integration accuracy in recurrent neural networks} 

\author{Yanlin Zhang}
\altaffiliation{School of physics and Electronic Engineering, Jiangsu University, Zhenjiang, Jiangsu 212013, China}  
\author{Yan Zhang}
\altaffiliation{Teaching Department, Jiangsu University Jingjiang College, Zhenjiang 212028, Jiangsu, China } 
\author{Muhua Zheng}
\altaffiliation{School of physics and Electronic Engineering, Jiangsu University, Zhenjiang, Jiangsu 212013, China} 
\author{Kesheng Xu}
\altaffiliation{School of physics and Electronic Engineering, Jiangsu University, Zhenjiang, Jiangsu 212013, China}  
\email{ksxu@ujs.edu.cn}

\date{\today}

\begin{abstract}
Experimental evidence indicates that intrinsic temporal dynamics operating across multiple time scales are closely associated with the emergence of periodic spatial activity of increasing complexity. However, how information encoded in grid-like firing patterns for path integration is processed across these intrinsic time scales remains unclear. To address this question, we introduce adaptive time scales through a leak term in recurrent neural networks (RNNs), forming leaky RNNs discretized from the continuous attractors of firing rate models. Our results demonstrate that leaky RNNs substantially enhance the emergence of well-defined and highly regular hexagonal firing patterns. Compared with vanilla RNNs lacking a leak term, the trained leaky RNNs produce more accurate position estimates while generating reliable grid-cell-like representations. Furthermore, under identical noise conditions, leaky RNNs consistently exhibit more stable dynamics and better-defined grid structures. The learned dynamics also give rise to stable torus attractors with a clear central hole, supporting robust and regular grid-like activity.	Overall, the dynamic leak acts as a low-pass filtering mechanism that protects recurrent neural circuitry from noise, stabilizes network dynamics, and improves path-integration accuracy in recurrent neural networks.	

\end{abstract}

\maketitle

\section{Introduction}
The hippocampal principal neurons involved in path-integration-based navigation have attracted significant attention in recent decades \cite{bush2014grid,morris2023chicken,birch2007rectangular,rebecca2025spatial,sorscher2019unified,dong2024grid,doeller2010evidence}. Numerous neuroscientific studies have demonstrated that different types of principal cells in the hippocampus contribute to the underlying neural mechanisms responsible for path-integration-based navigation. These principal cells include place cells \cite{o1979review}, head cells \cite{taube2007head}, grid cells \cite{hafting2005microstructure}, and other heterogeneous neurons (e.g., speed cells \cite{kropff2015speed} and boundary cells \cite{solstad2008representation,poulter2018neurobiology}), all of which have played a major role in advancing our understanding of spatial cognition and memory. Place cells in the hippocampus are typically active in specific locations within a given environment, supporting the cognitive map of known locations in rodents and episodic memory in humans. Head cells and grid cells, observed in the medial entorhinal cortex (MEC) and surrounding areas \cite{birch2007rectangular,rebecca2025spatial,sorscher2019unified}, encode spatial information differently. Head direction cells encode the animal's head direction in the horizontal plane independently of location, while grid cells have spatial firing fields that repeat periodically in a hexagonal pattern \cite{birch2007rectangular,rebecca2025spatial,sorscher2019unified,dong2024grid,doeller2010evidence}. Grid cells play key roles in coding space,
and integrating self-motion (cognitive map-based navigation and path integration) \cite{moser2008place,rowland2016ten,moser2014grid} and  is critical for  planning direct trajectories to goals (vector-based navigation)\cite{banino2018vector}. Finally, boundary cells fire when the animal reaches a specific distance or direction relative to environmental boundaries identified in the MEC. In this work, we primarily focus on the properties and functions of grid cells in encoding spatial information and supporting path integration through computational neuronal models.

Over the past two decades, mechanistic models of numerous continuous attractor networks (CAN)  have been proposed to elucidate the mechanisms underlying grid-like firing patterns \cite{dong2024grid, giocomo2011computational, burak2009accurate,fuhs2006spin}. Mechanistic models account for hexagonal firing fields as the result of pattern-forming dynamics in a recurrent neural network with hand-tuned center-surround connectivity\cite{NEURIPS2019_6e7d5d25}.An effective model of grid cells should capture several key features: the emergence of periodic spatial firing patterns, the robustness of this periodicity across varying running speeds and directions, the variability of spatial scales across the cell population, and the presence of temporally structured phenomena such as phase shifts or remapping in altered virtual environments\cite{giocomo2011computational}. Existing models that meet some of these criteria generally fall into two main categories. The first, mechanistic models, includes oscillatory-interference models \cite{giocomo2011computational, o2005dual} and continuous attractor neural (CAN) network  frameworks \cite{hafting2005microstructure, mcnaughton2006path, burak2009accurate}.
In oscillatory-interference models, the characteristic hexagonal firing pattern emerges from the interference of multiple membrane-potential oscillations whose frequencies are modulated by the animal’s running speed and direction \cite{giocomo2011computational, o2005dual}.
By contrast, CAN models generate hexagonal grid patterns through local network dynamics with center–surround connectivity \cite{GAORecurrent, dong2024grid}, enabling accurate and robust path integration of self-motion cues. Extensions of CAN models further incorporate synaptic plasticity\cite{widloski2014model} and plastic inputs from landmark cells\cite{ocko2018emergent}, allowing the integration of both self-motion and sensory landmark information\cite{campbell2018principles,GAORecurrent}. In a nutshell, mechanistic models account for hexagonal firing fields as the result of pattern-forming dynamics in a recurrent neural network with hand-tuned center surround connectivity\cite{sorscher2019unified}.

Popular discrete recurrent neural networks (RNN) constitute a second class of models, often referred to as normative models, which have emerged alongside advances in machine learning and deep learning techniques \cite{banino2018vector,kanitscheider2017training,sorscher2019unified,sorscher2023unified}. These models aim to uncover the mechanisms underlying the formation of the hexagonal firing patterns observed in grid cells, which are believed to be optimal for path integration. RNNs are typically trained using gradient-based methods, such as backpropagation through time. However, as the sequence length increases, the gradients of the loss function tend to decay exponentially, leading to the well-known vanishing gradient problem \cite{dey2017gate,chung2014empirical,salem2021gated}. To address long-term temporal dependencies, architectures such as long short-term memory (LSTM) RNNs and gated recurrent unit (GRU) RNNs \cite{cho2014learning} have been widely employed in path integration tasks. For example, LSTM RNNs can perform two-dimensional simultaneous localization and mapping tasks, although they do not exhibit explicit hexagonal grid patterns \cite{kanitscheider2017training}. RNNs trained via deep reinforcement learning have been shown to develop grid cell–like representations \cite{banino2018vector}. Overall, LSTM-, GRU-, and ReLU-based RNNs have proven effective in predicting the structure of emergent firing patterns.

RNNs enable machines to perform tasks involving sequential data, such as navigation, path integration, and dynamical system identification, without requiring explicit numerical integration of differential equations \cite{monfared2020transformation}. However, a common assumption underlying many discrete-time RNN formulations--when derived from their continuous-time counterparts--is the neglect of intrinsic time scales governing population activity \cite{monfared2020transformation}. Experimental evidence suggests that continuous-time formulations of neural systems are closely linked to biological constraints on intrinsic neuronal time scales\cite{monfared2020transformation,quax2020adaptive}. Neural responses can span a wide range of temporal scales: for instance, retinal responses to a flashing headlight occur on very short time scales \cite{quax2020adaptive}, whereas processes such as actor–critic learning in egocentric visual navigation unfold over much longer durations\cite{linemergence}. Moreover, continuous attractor neural network models with time scales are often highly nonlinear and potentially stiff \cite{jaeger2001echo,jaeger2007optimization}, which is usually smooth and eases the mathematical study of many neural phenomena, like those of periodic and non-periodic firing patterns for grid cells. Overall, a key limitation in deriving discrete-time RNNs as approximations of population firing-rate dynamics is the general neglect of neural time scales, which play a fundamental role in shaping neural activity across multiple temporal scales.

To date, the relationship between path integration and the emergence of hexagonal grid patterns under complex, biologically plausible constraints remains unclear. To address this fundamental issue, Sorscher et al. \cite{sorscher2023unified} proposed a unifying theory showing that vanilla RNNs with rectified linear unit (ReLU) activation functions, tend to produce more regular and well-defined hexagonal grid maps than LSTM-based architectures\cite{sorscher2023unified,fernandez2024grid}. In that framework, grid patterns arise from the learning dynamics of recurrent connectivity and can be interpreted as a form of symmetry-breaking or pattern formation. More importantly, their framework demonstrates that robust grid-like representations can emerge in normative models trained for path integration under two simple biological constraints: nonnegative firing rates \cite{dordek2016extracting} and a center–surround structure in inputs to place cells. Inspired by this unifying perspective, subsequent study has shown that vanilla RNNs composed of spiking neural units with biologically realistic architectures can support path integration and generate representations that more closely resemble the hexagonal grid patterns observed in the entorhinal–hippocampal system \cite{GAORecurrent}. However, the role of time scales -- particularly favoring slow dynamics capable of tracking continuous trajectories -- in the emergence of grid-like firing patterns in  ReLU-based RNNs remains an open question.

The presence of the leak term may function as a form of implicit velocity integration or positive feedback loop, encouraging slow dynamics that track continuous trajectories\cite{fuhs2006spin,burak2009accurate}, to facilitate the emergence of grid-like firing patterns.  In this work, we introduce adaptive time scales in recurrent neural networks, forming the leaky RNNs, which are discretized from the continuous-time dynamics of leaky integrator neurons \cite{jaeger2001echo,jaeger2007optimization,quax2020adaptive}.  Our results demonstrate that leaky RNN significantly enhances the formation of well-defined and highly regular hexagonal firing patterns, as confirmed by both Grid Score (GS) and Spatial autocorrelation (SAC) analyses.
Moreover, a comparison between predicted and true trajectories reveals that the trained leaky RNN provides more accurate position estimates, as measured by mean squared error (MSE), a crucial factor for generating reliable grid-cell-like firing patterns. Notably, the leaky RNN consistently outperforms RNNs exhibiting more stable and well-defined grid patterns even under identical noise conditions. Finally, we analyze the torus attractors obtained from both RNN and leaky RNN models, reinforcing the conclusion that leaky RNNs -- characterized by a visible hole at the center -- are more adept at producing stable and regular grid-like firing patterns compared to their RNN counterparts. The main findings suggest that the leakage  acts as a structural prior encouraging smooth temporal integration during path integration.

\section{Structure of the network}
\subsection{Recurrent Neural Network}
\subsubsection{Leaky Recurrent Neural Network}

The recurrent network consists of $N_r = 4096$ units, each receiving a two-dimensional body velocity input $\bm{V}^{t} = (v_x^t, v_y^t)$ at time $t$. Each neuron's firing rate \(r_i^t\) in conventional recurrent neural networks is computed using an activation function.Continuous-time recurrent neural networks consist of model neurons governed by the following general form:
		\begin{equation}
			\tau \dfrac{d\bm{r}}{dt} = -\bm{r} + \phi \left(\bm{W^{rec}r} + \bm{W^{in}V} \right)
			\label{CRA}
		\end{equation}
Where $\tau>0$ is a global time constant of the recurrent neural network, and $\phi(\cdot)$ denotes the activation function. $\bm{W}^{\mathrm{rec}}$ and $\bm{W}^{\mathrm{in}}$ are the recurrent and velocity-input weight matrices of sizes $(N_r, N_r)$ and $(N_r, 2)$, respectively. To incorporate explicit time scales into the RNN, after using the first-order Euler approximation with a time-discretization step $\delta$,we had:
		\begin{equation}
			\bm{r}^{t+1} = (1-\dfrac{\delta}{\tau})\bm{r}^t + \dfrac{\delta}{\tau}\phi \left(\bm{W^{rec}r}^{t} + \bm{W^{in}V}^{t} \right)
			\label{leakRNN0}
		\end{equation}

Using the substitution $\alpha = \delta/\tau$ in Eq.~\hyperref[leakRNN0]{\ref{leakRNN0}}, we obtain: 
	
			\begin{equation}
				\bm{r}^{t+1} = (1-\alpha)\bm{r}^t + \alpha\phi \left(\bm{W^{rec}r}^{t} + \bm{W^{in}V}^{t} \right)
				\label{leakRNN}
			\end{equation}
Which corresponds to leaky RNNs \cite{jaeger2001echo,jaeger2007optimization,driscoll2024flexible}. Here, $\alpha \in (0,1]$ denotes the leak rate of the reservoir, which controls the flow of information in the network state. When $\alpha$ is small, the dynamics are dominated by the previous state; when $\alpha = 1$, the model reduces to a vanilla RNN\cite{sorscher2023unified}, where the next reservoir state depends solely on the current state and the external input. In this work, we employ leaky RNN for both the path integration task and the training process. Each element of the input weight matrix $\bm{W}^{\text{in}}$ in the leaky RNN (Eq.~\hyperref[leakRNN]{\ref{leakRNN}}) is independently sampled from a uniform distribution, $W_{ij}^{\text{in}} \sim U\left(-\tfrac{1}{\sqrt{N_r}},\, \tfrac{1}{\sqrt{N_r}}\right)$.
\begin{figure}[ht]	
	\includegraphics[width=\linewidth]{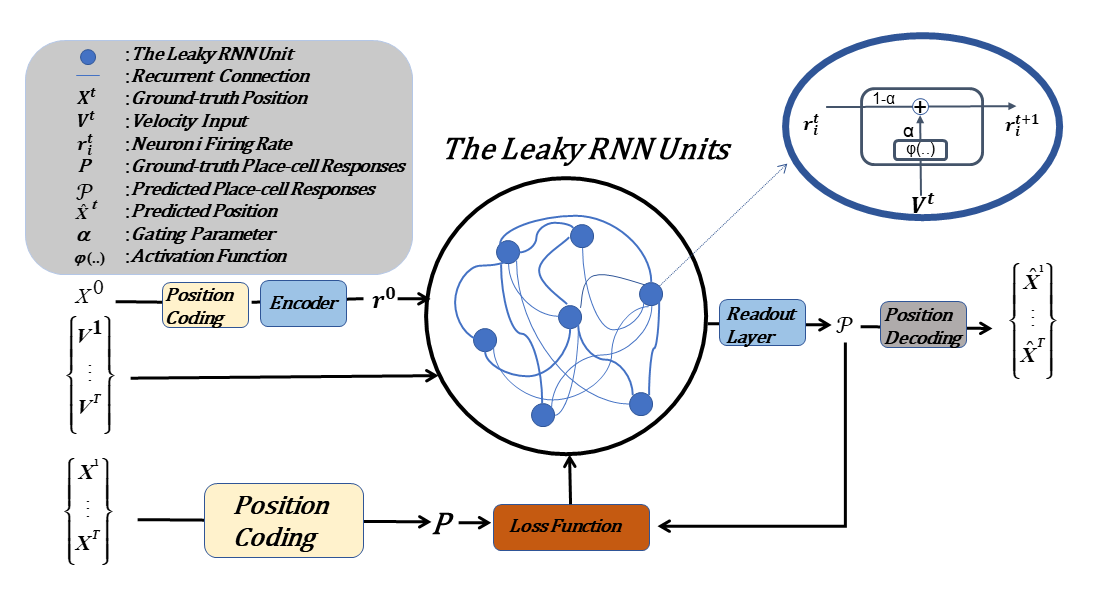}	  
	\caption{A schematic diagram of the path integration task and the training process. The leaky RNN is trained with adaptive time scales and ReLU activation functions, to predict the encoded integrated path from the initial position and the velocity sequence.}
	\label{fig1}
\end{figure}
\subsubsection{Read-out neurons of predicting the corresponding position sequence }
$W^{out}_{ij}$ is the strength of connection from $j_{th}$  to $i_{th}$ readout neuron. The readout matrix $W^{out}$ was initialized using the Xavier uniform initialization scheme \cite{glorot2010understanding}, with each element sampled from  $W^{out}_{ij} \sim U\left(-\sqrt{\tfrac{6}{N_r + N_p}},\; \sqrt{\tfrac{6}{N_r + N_p}}\right)$,  where $N_{p} = 512$ denotes the number of place cells in the population. This initialization maintains stable signal variance and balanced gradient propagation during both forward and backward passes, thereby enhancing training stability and convergence. The readout layer then computes the predicted place-cell responses as a weighted sum of recurrent neuronal activities through $W^{out}$:

\begin{equation}
	\mathcal{P}_{k}^t = \sum_{j=1}^{N_r} W^{out}_{k j} r_{j}^t,
	\qquad k \in \{1,\dots,N_p\},
	\label{eq:rnn_leak}
\end{equation}

\subsubsection{Trajectory Generation Process of Path integration task}

We trained leaky RNN on a path integration task within a supervised learning framework\cite{chaplot2018active, banino2018vector, GAORecurrent}. The network was provided with  encoded initial positions $\boldsymbol{X}^{0}$ and a sequence of velocity inputs $\boldsymbol{V}^{T}$, where $T$ represents the length of the trajectory sequences.The leaky RNN was trained to predict the corresponding encoded place-cell output sequence $\mathcal{P}_{k}^{t}$,then decoded it to reconstruct the integrated trajectory. The rat's trajectories were generated according to the motion scheme described below. For each simulated trajectory $b$, here $b$ denotes the index of a trajectory sample within a batch and the batch size set to 200,the sequences $\{\boldsymbol{X}_b^1,\cdots,\boldsymbol{X}_b^T\}$, where $\boldsymbol{X}_b^t=(x_b^t,\; y_b^t)\in\mathbb{R}^2$, and $\{\boldsymbol{V}_b^1,\cdots,\boldsymbol{V}_b^T\}$,where $\boldsymbol{V}_b^t=(v_{x,b}^t,\; v_{y,b}^t)\in\mathbb{R}^2$, denote the rat’s positions and velocities in the global coordinate frame, respectively. The rat’s initial position $\boldsymbol{X}_b^0$ and initial head direction $\theta_b^0$  were randomly initialized within a simulated $220\,\mathrm{cm}\times220\,\mathrm{cm}$ rectangular arena

At each time step, the rat's linear speed $||\boldsymbol{V}_b^t||$ is sampled from a Rayleigh distribution with scale parameter $b = 0.13 \times 2\pi$ $\mathrm{m/s}$, representing the magnitude of forward motion. Here, $0.13$ $\mathrm{m/s}$ serves as the base speed scale and provides a good fit to the statistical characteristics of real animal velocities \cite{banino_vectorbased_2018,sorscher_unified_2023}. The angular velocity $\omega_t$, for random changes in head direction,is sampled from a Gaussian distribution $\omega_t \sim \mathcal{N}(\mu,\sigma^2)$with the mean $\mu = 0$  and standard deviation $\sigma = 5.76\times2$, where 5.76 comes from a fitted experimental parameter for rotational speed\cite{raudies2012modeling}. When the rat approaches the boundary, an obstacle-avoidance mechanism is triggered, causing it to turn and decelerate to prevent escaping the box. Thus, the sequences of velocities in the global coordinate system and positions are obtained through:
\begin{subequations}
\begin{align}
\theta_b^t &= \theta_b^{t-1} + \delta_b^{t-1} + \omega_b^{t-1} \Delta t, \\
v_{x,b}^t &= \|\boldsymbol{V}_b^t\| \cos(\theta_b^t), \\
v_{y,b}^t &= \|\boldsymbol{V}_b^t\| \sin(\theta_b^t), \\
\boldsymbol{X}_b^t &= \boldsymbol{X}_b^{t-1} + \boldsymbol{V}_b^{t-1} \Delta t.
\end{align}
\label{EQ3}
\end{subequations}
$\|\bold{V}^t_{b}\|$ is the magnitude of the velocity vector at time $t \in [1,T]$ for the $bth$ trajectory. where $\theta_b^t$ denotes the head direction of the $b$-th trajectory at time step $t$, $\omega_b^t$ is the angular velocity randomly sampled for the $bth$ trajectory, and $\delta_b^t$ represents the additional turning angle induced by the boundary-avoidance mechanism. When the boundary-avoidance mechanism is not activated $\delta_b^t = 0$. In the supervised learning framework, the positions of the place cell centers $\{\bold{c}_{1},\cdots,\bold{c}_{N_{p}}\}$ with $\boldsymbol{c}_{k}\in \mathbb{R}^2$, which are uniformly and randomly sampled within the environment prior to training and remain fixed throughout the training process\cite{sorscher2023unified,GAORecurrent}. The corresponding encoded place-cell output sequence $P_{k}^{t}$, given the position $\boldsymbol{X}^t$, is defined as the difference between softmax tuning curves:

\begin{equation}
	\begin{split}
		P_{k}^{t}  & = \mathrm{softmax}\left(-\frac{\|\boldsymbol{X}^t - \boldsymbol{c}_{k}\|^2}{2\sigma_E^2}\right) - \mathrm{softmax}\left(-\frac{\|\boldsymbol{X}^t - \boldsymbol{c}_{k}\|^2}{2\sigma_I^2}\right)  \\
		  & = \frac{
		 	\exp\left(-\frac{\|\boldsymbol{X}^t - \boldsymbol{c}_{k}\|^2}{2\sigma_E^2}\right)}{
		 	\sum_{j=1}^{N_P} \exp\left(-\frac{\|\boldsymbol{X}^t - \boldsymbol{c}_{j}\|^2}{2\sigma_E^2}\right)
		 }
		 -
		 \frac{
		 	\exp\left(-\frac{\|\boldsymbol{X}^t - \boldsymbol{c}_{k}\|^2}{2\sigma_I^2}\right)
		 }{
		 	\sum_{j=1}^{N_P} \exp\left(-\frac{\|\boldsymbol{X}^t - \boldsymbol{c}_{j}\|^2}{2\sigma_I^2}\right)
		 }.
	\end{split}
\label{EQ1}
\end{equation}

Here, $\sigma_E = 20$ $ \rm{cm}$ and $\sigma_I = 40 $ $\rm{cm}$ denote the standard deviations (i.e., spatial widths) of the excitatory and inhibitory place-cell receptive fields, respectively.

\subsubsection{Noisy input}

To simulate imperfect sensory perception, we also introduce stochastic noise to corrupt the linear speed
$\|\mathbf{V}_b^t\|$, and then recompute the corresponding noisy velocity vectors as follows:

\begin{equation}
	\tilde{\mathbf{V}}_b^t = [\tilde{v}_{x,b}^t, \, \tilde{v}_{y,b}^t]
	= [\|\tilde{\mathbf{V}}_b^t\|\cos(\omega_t), \, \|\tilde{\mathbf{V}}_b^t\|\sin(\omega_t)].
\end{equation}

where $\|\tilde{\mathbf{V}}_b^t\|$ represents the noisy linear speed after perturbation. These noisy velocity sequences are then used as inputs to the leaky RNN, enabling the network to learn how to infer the true trajectories from the corrupted motion signals. 
Specifically, two types of noise are considered: Gaussian white noise and Ornstein–Uhlenbeck (OU) noise, which model different forms of sensory uncertainty encountered in animal navigation. Gaussian white noise represents instantaneous, uncorrelated perturbations in the motion input, effectively simulating instantaneous uncertainty in velocity estimation. After applying this noise, the linear speed becomes:
\begin{equation}
	\|\tilde{\mathbf{V}}_b^t\|_{\text{gauss}} = \|\mathbf{V}_b^t\| + h\,\eta_b^t,
	\quad ,
	\label{eq:gaussian_noise}
\end{equation}

where $\eta_b^t \sim \mathcal{N}(0, 1)$ and $h$ controls the overall noise intensity.The Ornstein–Uhlenbeck (OU) process has long been an effective model for synaptic noise, which is governed by:
\begin{equation}
	\frac{d\xi_b^t}{dt} = -\frac{1}{\tau}\xi_b^t + \sigma \sqrt{\frac{2}{\tau}}\,\mathcal{N}(0,1),
	\label{eq:ou_process}
\end{equation}

where $\xi_b^t$ is the OU stochastic process, $\tau$ controls the correlation timescale, and $\sigma$ determines the noise amplitude. After adding OU noise, the noisy linear speed becomes:

\begin{equation}
	\|\tilde{\mathbf{V}}_b^t\|_{\text{OU}} = \|\mathbf{V}_b^t\| + \xi_b^t,
	\label{eq:ou_velocity}
\end{equation}

This process maintains smooth fluctuations and slow drift, closely resembling the cumulative sensory bias observed during self-motion estimation in biological path integration.

\subsubsection{Learning rule of loss function  utilizing the cross-entropy function}

The learning rules in this work consist of the cross-entropy function \cite{connor2024correlations} and a regularization term with a small penalty $L_{w}$\cite{sorscher2023unified}. The regularization term is defined as   $L_{w} = \lambda \| W_{\text{rec}} \|^2$, which helps prevent the model from over fitting to specific features and improves its generalization capability. Here, $\lambda$ is the regularization coefficient that controls the strength of the $L_w$ penalty. The learning rule for the loss function is then given as follows:
\begin{equation}
			\mathcal{L} = - P^{\top} \ln\mathcal{P} + \lambda \, \| W_{\text{rec}} \|^2
			\label{eq:loss_function}
\end{equation}
\subsection{Statistical methods to capture hexagonal grid patterns  }
\subsubsection{Spatial Autocorrelogram (SAC) and Grid score (GS)}
\begin{figure}[ht]	
	\includegraphics[width=\linewidth]{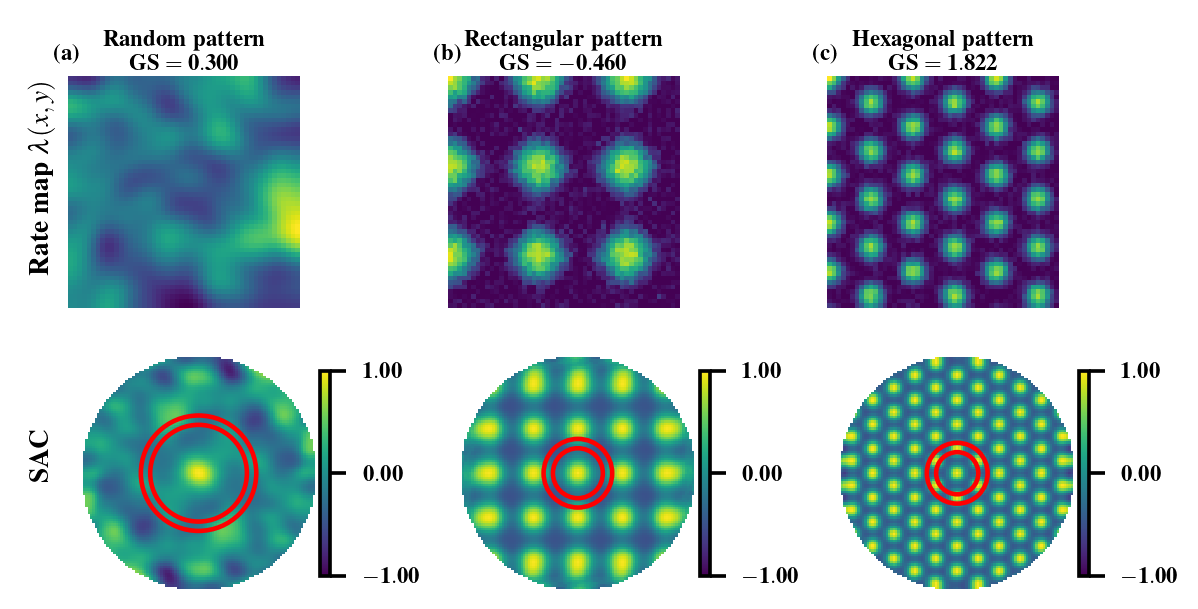}	  
	\caption{Three representative firing patterns (top panels) are quantified using spatial autocorrelograms (SAC; bottom panels): (a) random firing pattern, (b) rectangular firing pattern, and (c) hexagonal firing pattern.}
	\label{fig2}
\end{figure}

\textbf{Grid cell spatial firing rate map:} Rate maps were constructed for arena and track recordings by sorting the position data  within a partitioned bins.  The grid cell rate maps (e.g., in Fig.~\hyperref[fig2]{\ref{fig2}(a)}) were computed as follows.  Simulated rats performed path integration tasks with arena size (AZ) $220\text{cm}\times 220\text{cm}$.   To estimate the averaged instantaneous firing activity $r_i^t$ at different locations, the AZ box ($220 \text{cm}\times 220 \text{cm}$) digitized into small spatial bins of $4.4 \text{cm} \times 4.4 \text{cm}$. The firing rates  $r_i^t$ for each visited small spatial bin of the same location were time--averaged, while the rates for unvisited bins were set to zero. Note that, in this work, we obtain $N=4096$ rate maps for different neurons. Therefore, the rate map $\lambda_i(x,y)$ of neuron $i$ at the spatial location $(x,y)$ is defined as:

\begin{equation}
	\lambda_i(x,y)=
	\frac{1}{T}\int_{0}^{T} r_i^t(x,y)dt,
	\label{eq:ratemap}
\end{equation}
Where $T$ denotes the trial duration of the simulated rat’s trajectory as before.
	
\textbf{Spatial autocorrelograms (SAC)} of rate maps were used to assess the periodicity, regularity, and orientation of cells with multiple firing fields in open fields\cite{sorscher2023unified}. SAC were based on Pearson’s product moment correlation coefficient with corrections for edge effects and unvisited locations.  With $\lambda(x,y)$ denoting the average rate of a cell at location $(x, y)$, the SAC between the original rate map at location $(x,y)$ and spatial lags of $(\tau_x,\tau_y)$ was estimated as\cite{JUN20201095}:

\begin{equation}
r(\tau_x,\tau_y)=\frac{n\sum \lambda(x,y)\lambda(x-\tau_x,y-\tau_y)
-\sum \lambda(x,y)\sum \lambda(x-\tau_x,y-\tau_y)}{
\sqrt{
n\sum \lambda(x,y)^2-\left(\sum \lambda(x,y)\right)^2
}
\sqrt{
n\sum \lambda(x-\tau_x,y-\tau_y)^2-\left(\sum \lambda(x-\tau_x,y-\tau_y)\right)^2
}
},
\label{eqSAC}
\end{equation}
where the summation is taken over all $n$ spatial bins for which values are defined in both $\lambda(x,y)$ and the shifted map $\lambda(x-\tau_x,y-\tau_y)$. Equation~\hyperref[eqSAC]{\ref{eqSAC}} thus effectively accounts for the reduced overlap near the boundaries. Specifically, for an ideal hexagonal grid pattern, the SAC exhibits a prominent central peak surrounded by six evenly spaced secondary peaks, reflecting the intrinsic periodicity of the underlying spatial representation.

\textbf{Grid scores}, which quantify the degree of spatial periodicity, was calculated for each recorded cell by analyzing by taking a circular sample of the spatial autocorrelogram, centered on the central peak but with the central peak excluded, and comparing rotated versions of this sample\cite{langston2010development,perez2016visual}.
 Pearson correlation coefficients $r_{\theta}$ were calculated between the selected circular region (the red annular mask in Fig.~\hyperref[fig2]{\ref{fig2}(b)}) of the matrix $r(\tau_x,\tau_y)$ and its rotated versions of itself at angles $\theta = 30^\circ, 60^\circ, 90^\circ, 120^\circ,$ and $150^\circ$. The rotations at $60^\circ$ and $120^\circ$ formed one group, while those at $30^\circ$, $90^\circ$, and $150^\circ$ formed the other. The grid score was then defined as:
	\begin{equation}
		GS=\frac{r_{60^\circ}+r_{120^\circ}}{2}-\frac{r_{30^\circ}+r_{90^\circ}+r_{150^\circ}}{3}.
		\label{eqGS}
	\end{equation}
This procedure was repeated 10 times for each neuron to obtain surrogate distributions. Since the Pearson correlation coefficients at all rotation angles lie within the interval $[-1,1]$, the theoretical range of the grid score (GS), as defined in Eq.~\hyperref[eqGS]{\ref{eqGS}}, is $[-2,2]$. A value of $GS = 2$ corresponds to an ideal hexagonal grid structure, whereas $GS \rightarrow 0$ indicates weak or indistinct sixfold symmetry.Values of $GS \leq 0$ suggest that the firing pattern lacks a regular grid-like structure. Overall, the GS provides a quantitative measure of the periodicity and rotational symmetry of the firing patterns, whereas the SAC offers detailed insight into the spatial organization and regularity of the firing fields.
	
\subsubsection{Mean Squared Error (MSE)}

For a trajectory of length $T$,  let $\mathbf{X}^t = (x^t,y^t)$  and $\hat{\mathbf{X}}^t = (\hat{x}^t,\hat{y}^t)$ for $t = 1,\dots,T$ denote the ground-truth 2D position and the predicted position at time $t$, respectively.  The step-wise position error is therefore defined as
\begin{equation}
e_t = \sqrt{\left(\hat{x}^t - x^t \right)^{2}
    +
    \left(\hat{y}^t - y^t \right)^{2}
  }.
\end{equation}
which quantifies the instantaneous deviation between the predicted and true positions at each time step. The time-averaged position error for the $ith$ trajectory is computed as $E^{i} = \frac{1}{T} \sum_{t=1}^{T} e_t $. Accordingly, the mean squared error (MSE) is defined as the population average of the step-wise position errors across all trajectories:
\begin{equation}
\text{MSE} =\frac{1}{M}
  \sum_{i=1}^{M} E^{(i)},
  \label{mse}
\end{equation}
where $M$ denotes the total number of trajectories.

\subsubsection{Persistent Homology}
	
To better reveal stable toroidal attractors characterized by a clear central hole and robust, regular grid-like activity, we employed persistent homology, a topological data analysis framework that extracts geometric features across multiple spatial scales. Persistent homology characterizes the multiscale structure of data by tracking topological invariants, including connected components ($H_0$), loop structures ($H_1$), and higher-dimensional cavities ($H_2$). Persistent homology records the birth and death scales of these features, filtering noise by defining persistence (lifetime) as the difference between death and birth scales, keeping only the most significant features\cite{zomorodian_computing_2005,ghrist_barcodes_2007,edelsbrunner_topological_2000}.
		
Persistent homology can be described by a persistence diagram or a persistence barcode\cite{obayashi2023stable,berry2020functional}, whose construction depends on the filtration defined over the dataset. In our analysis, central regions of neuronal rate maps were first extracted to minimize boundary artifacts. Population activity across spatial bins was then organized into a high-dimensional point cloud after removing invalid or nearly constant dimensions. The data were subsequently reduced to seven dimensions via principal component analysis. Cosine distances between samples were used to compute $H_0$, $H_1$, and $H_2$ persistence diagrams using Ripser\cite{ctralie2018ripser,Bauer2021Ripser}. Feature lifetimes were quantified, and independently shuffled point clouds served as background controls to evaluate the significance and stability of connected components, loops, and higher-dimensional cavities.

\begin{figure}[ht]	
	\includegraphics[width=\linewidth]{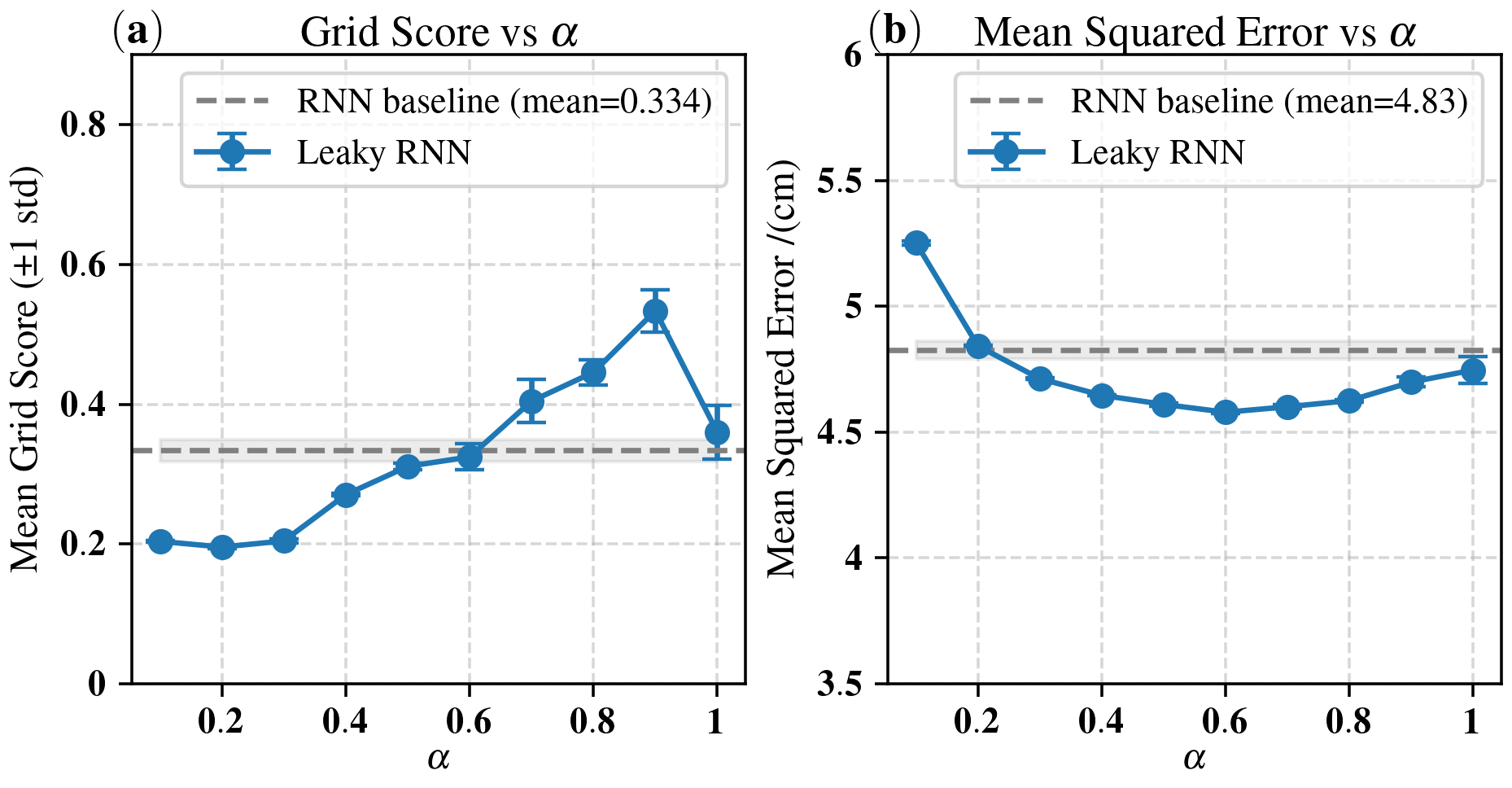}	  
	\caption{Statistical analysis of the grid score (a) and mean squared error (b) as a function of the leak  parameter $\alpha$, compared to the results from the standard RNN, emphasizing the impact of the leak mechanism on model performance.}
	\label{fig3}
\end{figure}

\begin{figure}[ht]	
	\includegraphics[width=\linewidth]{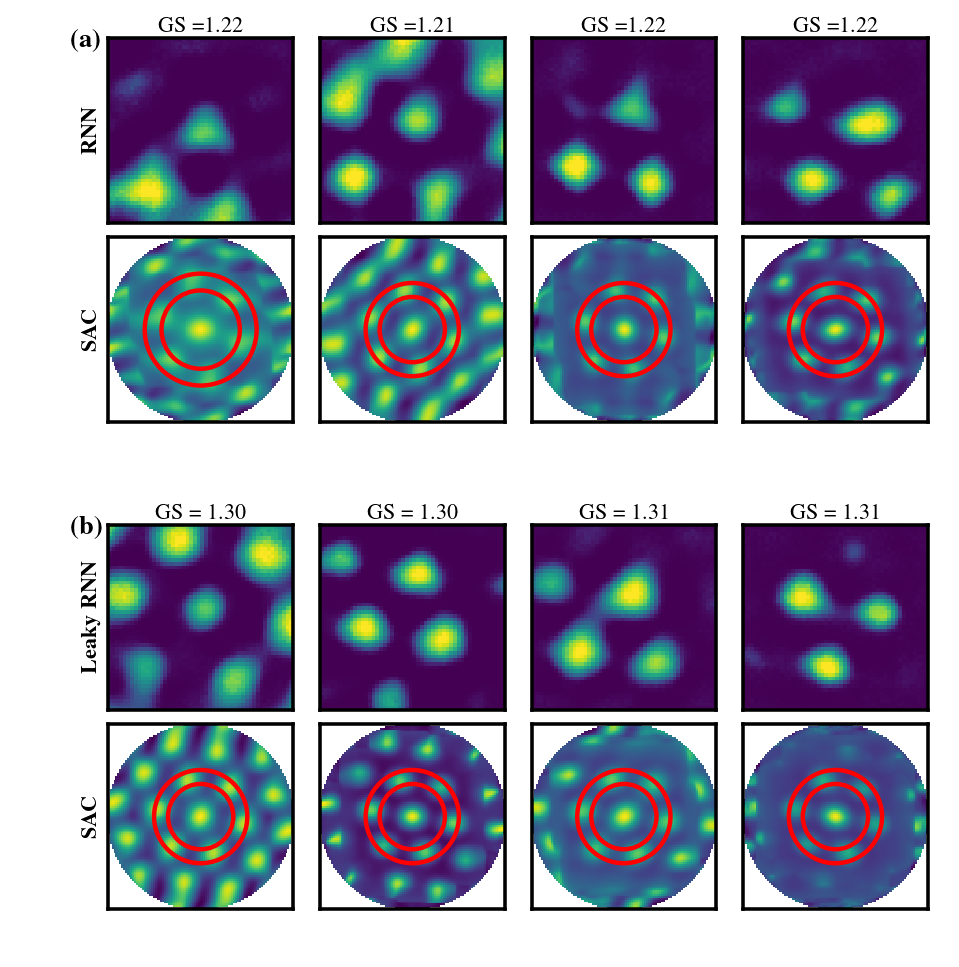}	  
	\caption{Comparison of the hexagonal firing patterns for grid cells obtained from the RNN models (a) and the leaky RNN models (b) with $\alpha = 0.95$. The patterns are quantified by Grid scores (shown at the top) and SAC (shown in the bottom panels of subplots a and b).}
	\label{fig4}
\end{figure}

\section{Results}

\subsubsection{The effect of introducing the leak term $\alpha$ on the emergence of grid-cell firing patterns in RNNs}

To assess the impact of the leak parameter $\alpha$ on the emergence of representative firing patterns, we compare the activity patterns generated by leaky RNNs with those produced by vanilla RNNs (i.e., $\alpha = 1$). The degree to which grid-like firing patterns emerge is quantitatively evaluated using the grid score and the mean squared error (MSE; Eq.~\hyperref[mse]{\ref{mse}}) across different values of the leak parameter $\alpha$ in both network types.

As shown in Fig.~\hyperref[fig3]{\ref{fig3}(a)}, when $\alpha \le 0.6$, the grid scores of the leaky RNN remain below the vanilla RNN baseline ($\mathrm{GS} = 0.334$), indicating that small values of $\alpha$ do not effectively support the emergence of grid-cell firing patterns. For $\alpha \in (0.6, 1.0)$, the grid scores exceed this baseline, revealing a range of leak parameters that more strongly promotes the formation of hexagonal firing structures. Fig.~\hyperref[fig3]{\ref{fig3}(b)} further shows that the MSE initially decreases below the vanilla RNN baseline ($\mathrm{MSE} = 4.83$) and then gradually increases toward the baseline as $\alpha$ grows. Overall, the results in Fig.~\hyperref[fig3]{\ref{fig3}} demonstrate that the leak parameter $\alpha$, which governs the intrinsic network dynamics, plays a crucial role in facilitating the emergence of grid-cell firing patterns. Both the grid score and MSE indicate the existence of an optimal intermediate range of $\alpha$.

We further investigate the development of hexagonal firing patterns in both the vanilla RNN and the leaky RNN by analyzing the grid score (GS; top panels in Figs.~\hyperref[fig4]{\ref{fig4}(a),(b)}) and the spatial autocorrelogram (SAC; bottom panels in Figs.~\hyperref[fig4]{\ref{fig4}(a),(b)}) for the case $\alpha = 0.95$. Across multiple representative examples, the leaky RNN consistently achieves higher grid scores than the vanilla RNN, indicating that the introduction of the leak parameter $\alpha$ facilitates the emergence of grid-like firing activity, leading to enhanced periodicity and more coherent alignment of firing fields.

Furthermore, the SAC results shown in Fig.~\hyperref[fig4]{\ref{fig4}} reveal that the hexagonal firing patterns observed in the leaky RNN are not only more pronounced but also exhibit substantially greater regularity and symmetry than those generated by the vanilla RNN. The SACs display clearer and more distinct hexagonal structures, indicating improved resolution and precision of the detected firing fields. Together, the GS and SAC analyses provide strong evidence that introducing the leak parameter $\alpha$ into the RNN facilitates the formation of well-defined and highly regular hexagonal firing patterns. The observed improvements in both periodicity (captured by the GS) and spatial organization (revealed by the SAC) highlight the critical role of leak-mediated dynamics in optimizing the representation of grid-like neural activity.

\subsubsection{Comparison of Integrated Navigation Paths Generated by the vanilla RNN and the leaky RNN}

\begin{figure}[ht]	
	\includegraphics[width=\linewidth]{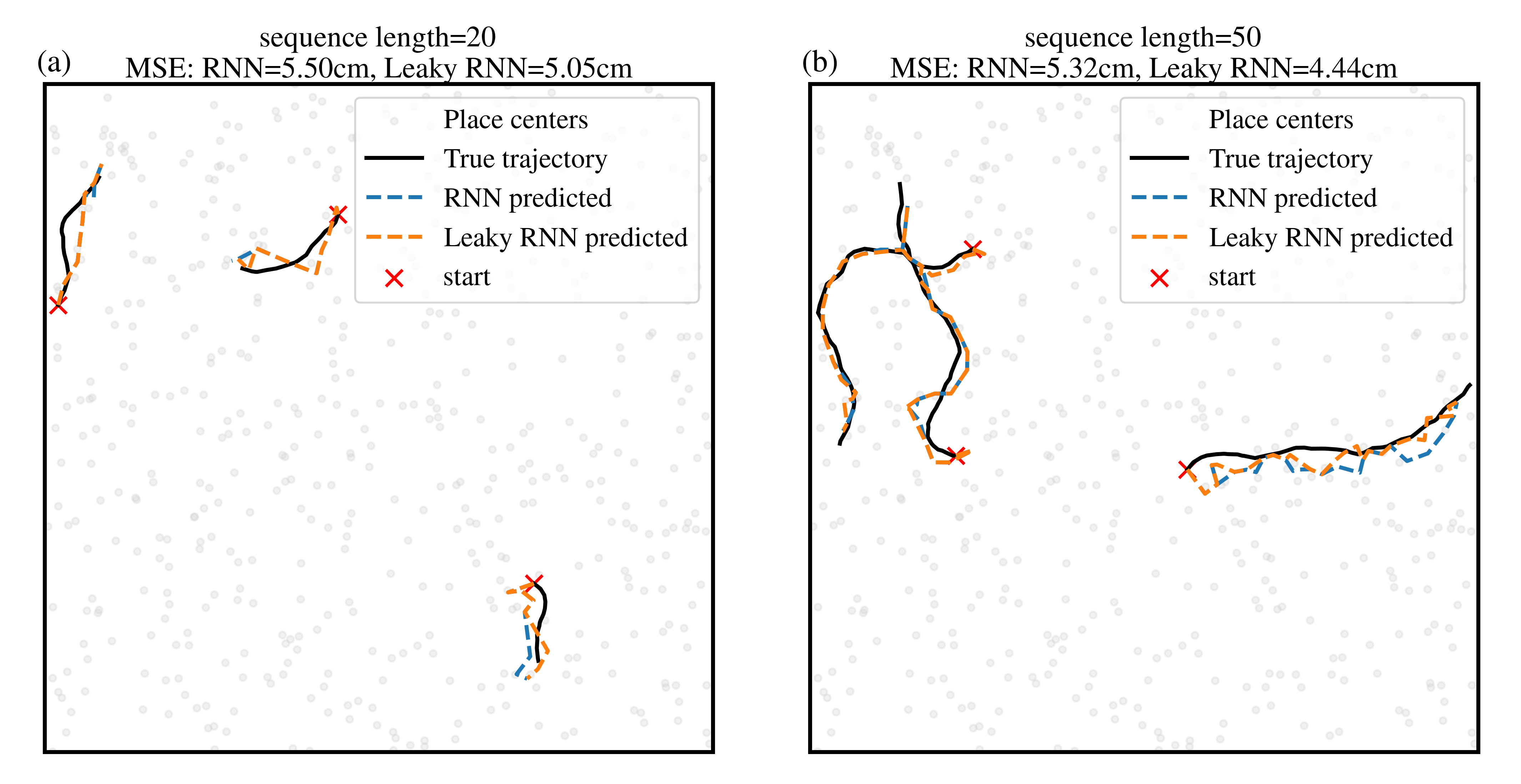}	  
	\caption{Illustration of the path integration task and training process, compared to the ground truth in the simulated environment, for the RNN and leaky RNN models with $\alpha = 0.95$. (a) Testing results for a sequence length of $T = 20$. (b) Testing results for an extended sequence length of $T = 50$.}
	\label{fig5}
\end{figure}

Path-integration accuracy is quantified using MSE between the predicted and ground-truth positions along the entire trajectory, similar approaches as in previous studies \cite{sorscher2023unified, GAORecurrent}. This metric assesses how closely the model’s predicted path matches the true trajectory training by different types of recurrent neural networks. Fig.~\hyperref[fig5]{\ref{fig5}} presents the performance of the trained leaky RNN and vanilla RNN models on the path-integration task, with their predicted trajectories compared directly against the ground truth. For visualization, the models are evaluated on three randomly selected trajectories with sequence lengths of $T = 20$ [Fig.~\hyperref[fig5]{\ref{fig5}(a)}] and $T = 50$ [Fig.~\hyperref[fig5]{\ref{fig5}(b)}], respectively, illustrating the models’ ability to generalize across different path lengths. For sequences of length $T = 20$, the leaky RNN achieves a slightly lower MSE between the predicted and true positions ($\mathrm{MSE} = 5.05$ cm) than the vanilla RNN ($\mathrm{MSE} = 5.50$ cm). A similar trend is observed for longer sequences ($T = 50$), where the leaky RNN again yields a reduced MSE ($\mathrm{MSE} = 4.44$ cm). These results indicate that the leaky RNN provides more accurate position estimates, which is crucial for supporting stable and reliable grid-cell–like firing representations.

\begin{figure}[t]
    \includegraphics[width=\linewidth]{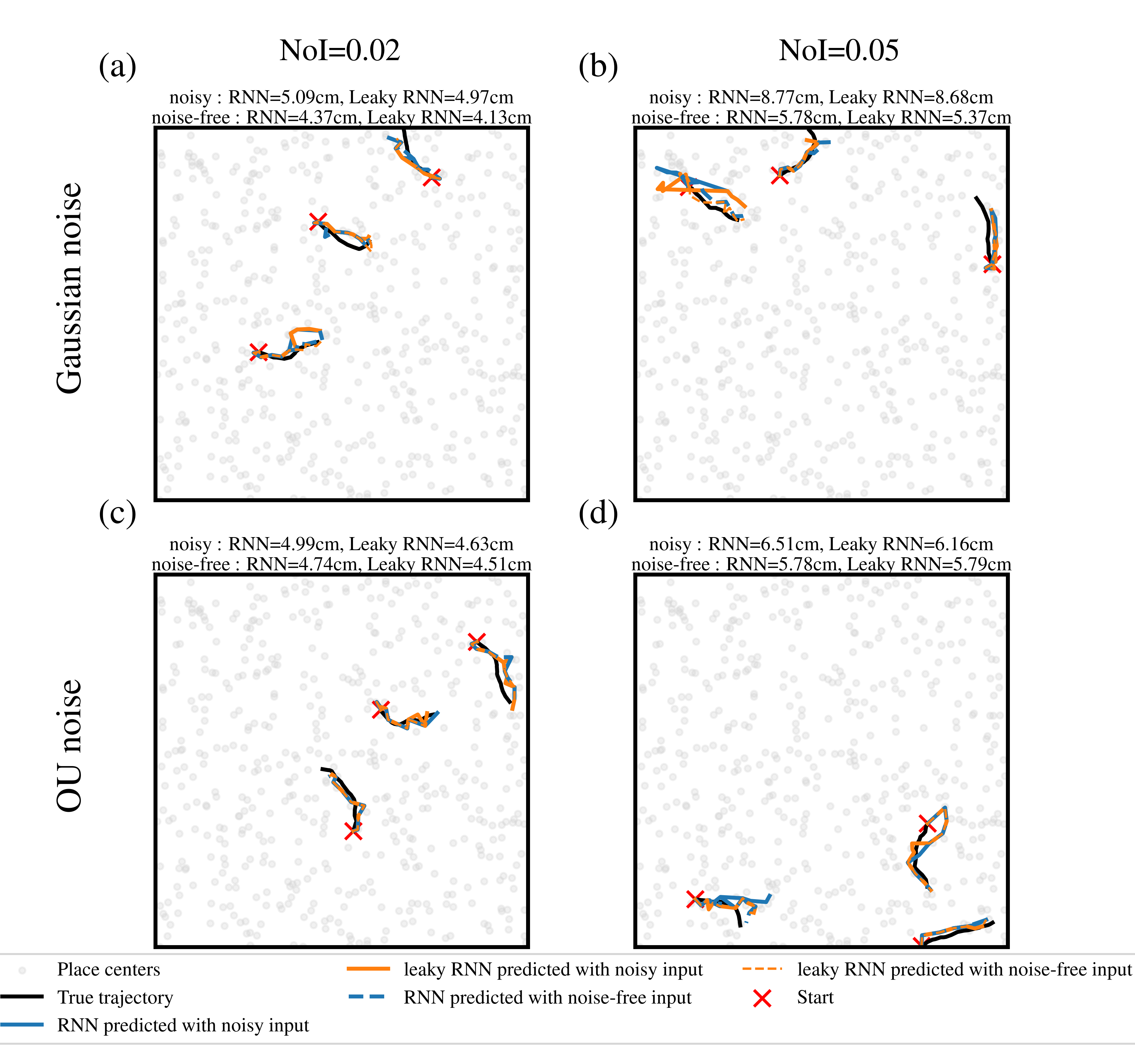}		  
	\caption{Comparison of predicted trajectories between the vanilla RNN and leaky RNN under Ornstein–Uhlenbeck noise and Gaussian noise conditions. Noise intensity: (a), (c) $\mathrm{NoI}=0.02$; (b), (d) $\mathrm{NoI}=0.05$.}
	\label{fig6}
\end{figure}

So far, all results presented have been obtained using the leaky RNN under noise-free conditions. To further assess its robustness and precision in position predictions, we introduce two types of stochastic perturbations to the input velocity signals: Gaussian white noise and Ornstein-Uhlenbeck (OU) noise. These noise models capture distinct forms of sensory uncertainty often encountered in biological navigation systems. By comparing the ground-truth position trajectories under noise-free conditions with those affected by each noise type, we try to evaluate how the leaky RNN responds to input perturbations and how its prediction accuracy changes relative to the noise-free baseline.

To evaluate robustness under noisy conditions, Fig.~\hyperref[fig6]{\ref{fig6}} compares predicted and ground-truth trajectories for sequences of length $T = 20$ under both Gaussian and Ornstein–Uhlenbeck (OU) noise. Under Gaussian noise [Figs.~\hyperref[fig6]{\ref{fig6}(a),(b)}], the leaky RNN consistently achieves lower MSE than the vanilla RNN and preserves this performance advantage as noise intensity increases. A comparable improvement is observed under OU noise [Figs.~\hyperref[fig6]{\ref{fig6}(c),(d)}], where the leaky RNN maintains reduced positional error across all tested noise levels. Together, these results demonstrate that incorporating leak dynamics substantially enhances the robustness and reliability of path integration, enabling stable trajectory estimation even in noisy environments.

\subsubsection{Impact of noise on grid cell firing pattern formation in the leaky RNN} 

\begin{figure}[ht]	
	\includegraphics[width=\linewidth]{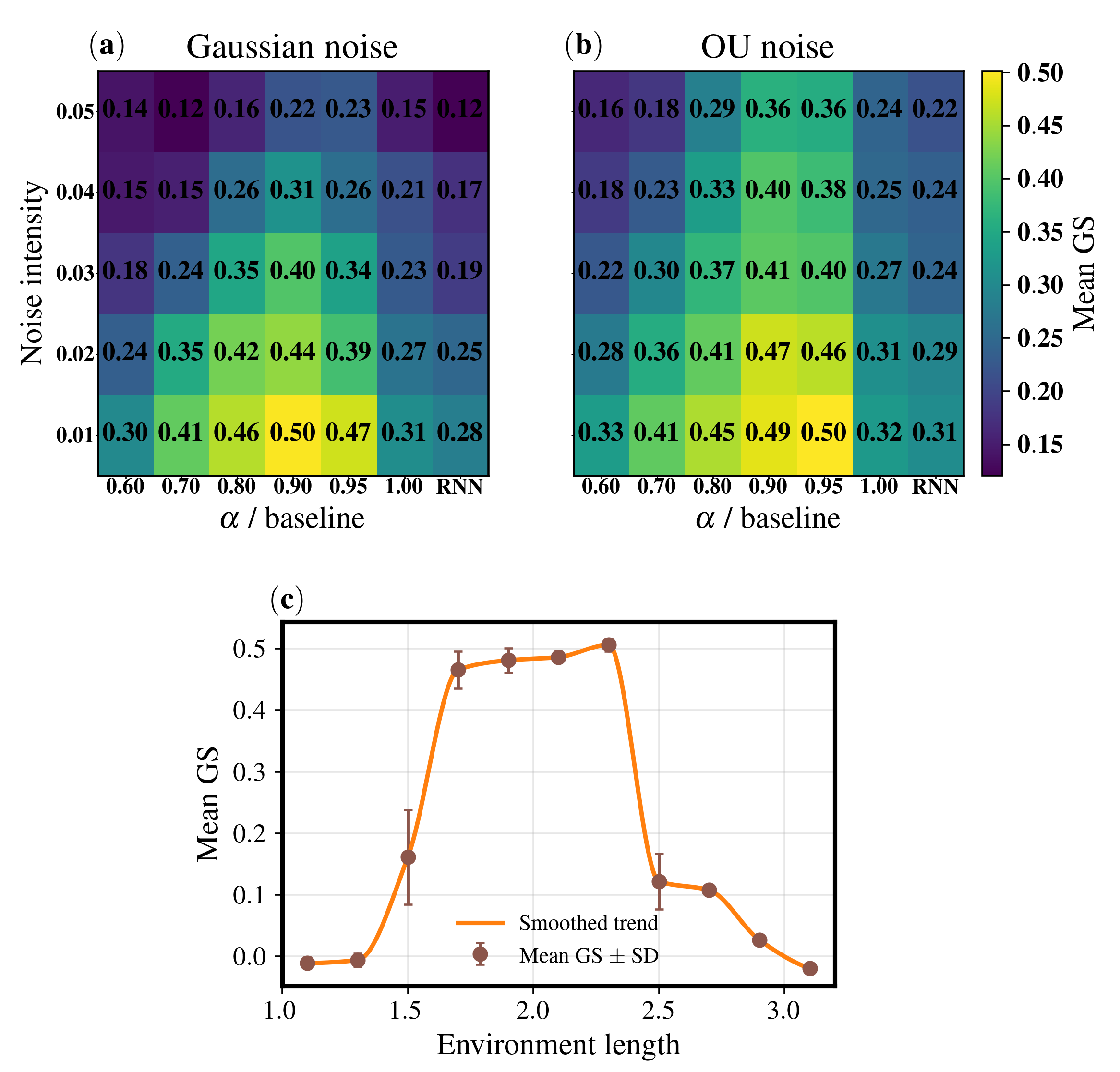}	  
	\caption{Combined effects of the leak parameter $\alpha$ and noise intensity on the emergence of grid-like firing patterns in the leaky RNN model. Results are shown under Gaussian white noise (a) and Ornstein–Uhlenbeck noise (b), with comparisons to the vanilla RNN model. (c) Mean grid score across different environment lengths at $\mathrm{NoI}=0.01$ and $\alpha=0.9$.}
	\label{fig7}
\end{figure}

To better understand the impact of noise on grid cell firing pattern formation in the leaky RNN, we analyze how the mean grid score varies as a function of both the leak parameter $\alpha$ and noise intensity. Specifically, we investigate the effects of two types of noise: Gaussian white noise and OU noise. The results, showing how grid cell patterns are influenced by these factors, are presented in Fig.~\hyperref[fig7]{\ref{fig7}}.

Specifically, when the noise intensity (NoI) is fixed, using $\text{NoI} = 0.01$ as an example, Fig.~\hyperref[fig7]{\ref{fig7}(a)} shows that the mean grid score in the leaky RNN gradually increases to a peak of 0.50 (indicated by yellow) as $\alpha$ increases, before decreasing to 0.31 (indicated by green) at higher values of $\alpha$. This is compared to the mean grid score obtained from the vanilla RNN under Gaussian noise (the right column, labeled 'RNN'). A similar analysis of the leaky RNN and RNN under OU noise is presented in Fig.~\hyperref[fig7]{\ref{fig7}(b)}, where the same trends are observed. Notably, the maximum grid score consistently occurs when $\alpha$ is in the range of (0.90, 0.95), and this peak remains robust across different noise strengths for both Gaussian and OU noise conditions. Furthermore, the mean grid score shows a gradual decline as noise strength increases, with this decrease observed across various levels of $\alpha$ for both types of noise.

From Figs.~\hyperref[fig7]{\ref{fig7}(a),(b)}, a well-defined optimal leak parameter $\alpha$ emerges under both Gaussian and OU noise conditions, suggesting that grid formation critically depends on an appropriate balance between memory retention and dynamic updating in recurrent activity. This optimal leak likely reflects intrinsic temporal constraints of internal neural dynamics interacting with environmental and spatial statistics\cite{quax2020adaptive,tani2016exploring,regazzoni2024learning} (see the Discussion section for further details). Figure~\hyperref[fig7]{\ref{fig7}(c)} further reveals how the mean grid score varies with increasing environment length at $\alpha = 0.9$. The results indicate that reliable grid-like firing patterns emerge only within an intermediate spatial scale: environments that are too small fail to support sufficient periodic structure, whereas excessively large environments degrade coherent spatial coding. The maximal Mean GS observed around an environment length of approximately $200$ cm. Collectively, these findings suggest that the leaky RNN provides a dynamical regime that stabilizes grid representations against noise, enabling more reliable path integration and demonstrating substantially greater robustness than the vanilla RNN.

\begin{figure}[htp]	
	\includegraphics[width=\linewidth]{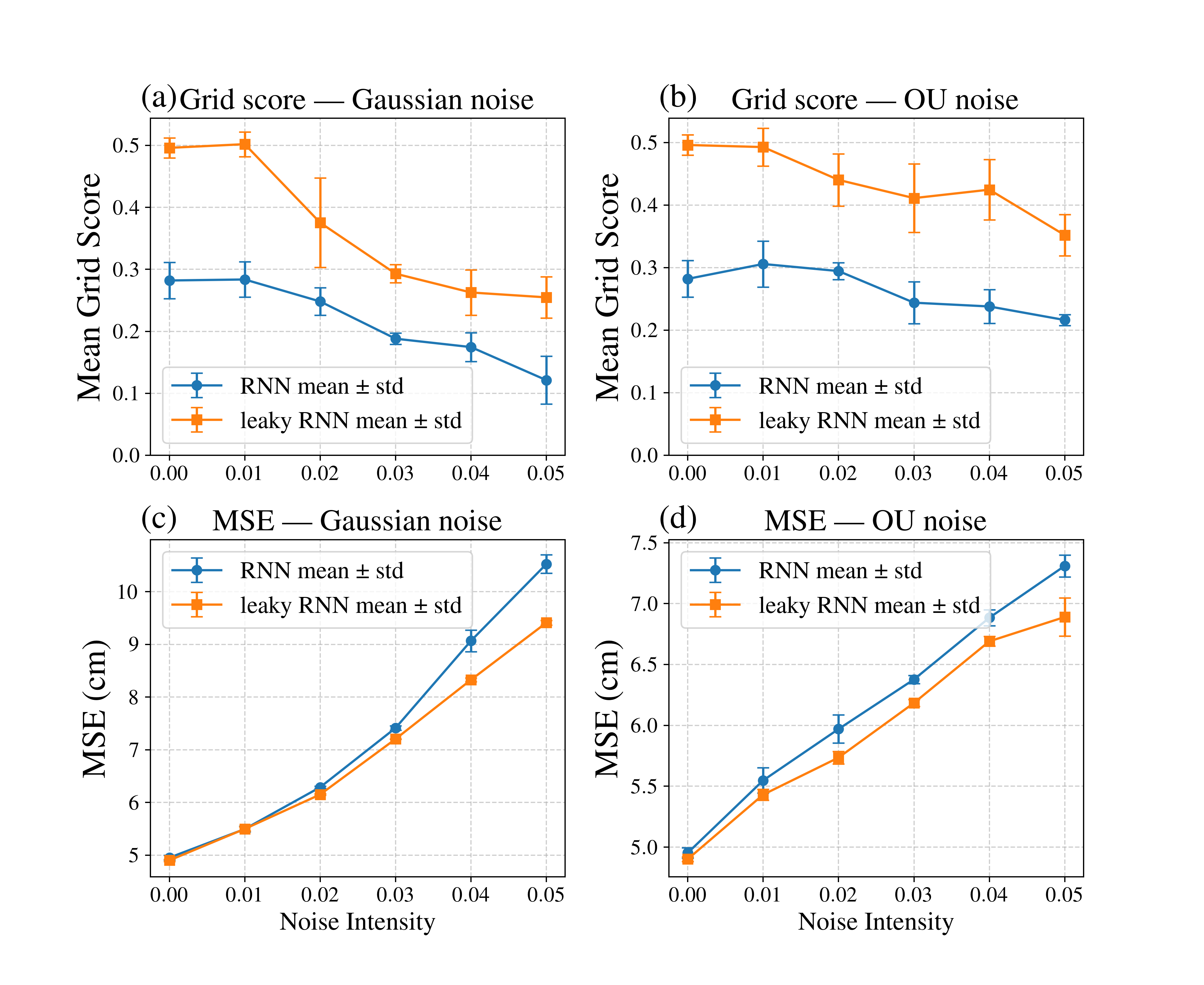}	  
	\caption{The grid scores (a,b) and mean squared errors (MSE,c,d) as functions of noise intensity for $\alpha = 0.95$ are evaluated under both Gaussian white noise and Ornstein–Uhlenbeck (OU) noise, highlighting the effects of noise intensity on the emergence of grid-like firing patterns.}
	\label{fig8}
\end{figure}

\begin{figure}[htp]	
	\includegraphics[width=\linewidth]{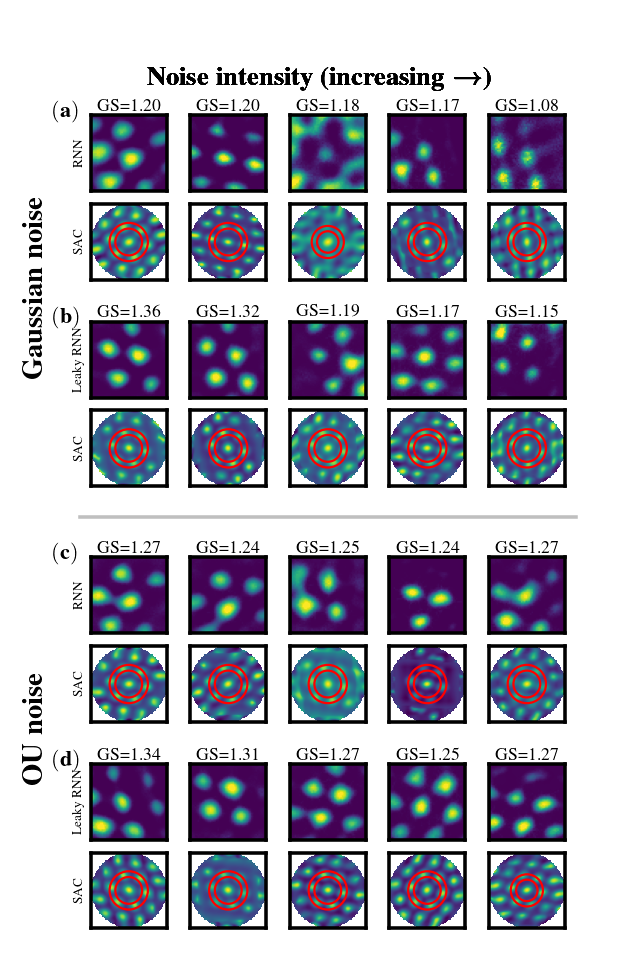}	  
	\caption{Comparison of hexagonal firing patterns in the vanilla RNN and leaky RNN models under Gaussian white noise (a, b) and Ornstein–Uhlenbeck noise (c, d), evaluated using Grid Score and Spatial Autocorrelation.}
	\label{fig9}
\end{figure}

To further investigate how noise affects the formation and stability of grid-like firing patterns, Fig.~\hyperref[fig8]{\ref{fig8}} presents the grid scores and  mean squared errors of the leaky RNN across a range of noise intensities under both Gaussian noise [Figs.~\hyperref[fig8]{\ref{fig8}(a),(c)}] and Ornstein–Uhlenbeck (OU) noise [Figs.~\hyperref[fig8]{\ref{fig8}(b),(d)}], together with the corresponding results from vanilla RNNs with $\alpha = 0.95$. As shown in Figs.~\hyperref[fig8]{\ref{fig8}(a),(b)}, the leaky RNN (orange curves) consistently achieves higher grid scores than the vanilla RNN (blue curves) across all noise intensity levels ($\text{NoI}$), irrespective of noise type. Although grid scores in both models gradually decline with increasing noise intensity, reflecting the disruptive effect of noise on pattern regularity, the leaky RNN preserves markedly stronger grid-like structure, maintaining a robust advantage throughout the entire noise range.

This observation is further corroborated by the MSE results in {Figs.~\hyperref[fig8]{\ref{fig8}(c),(d)}.} As the noise intensity increases, the the MSE for both models rise sharply, indicating reduced path-integration accuracy. At low noise levels, the RNNs exhibit coding errors comparable to those of the leaky RNN, suggesting that both models perform similarly when sensory uncertainty is minimal. However, as $\text{NoI}$ continues to increase, the the MSEs of vanilla RNNs escalate more rapidly and eventually exceed those of the leaky RNN. This divergence highlights the superior robustness of the leaky RNN: it not only maintains more regular grid-like firing patterns under moderate to strong noise, but also sustains lower MSEs , demonstrating greater resilience in position-estimation accuracy under noisy conditions.

The results from both the Grid score and MSE analyses clearly demonstrate that grid-like firing patterns in the leaky RNN can be reliably captured, even in the presence of noise. To further support this, Fig.~\hyperref[fig9]{\ref{fig9}} presents the grid patterns detected through the Grid score and SAC for both the RNN and leaky RNN under Gaussian and OU noise perturbations. Specifically, Figs.~\hyperref[fig9]{\ref{fig9}(a,b)} illustrate the evolution of grid firing patterns in the top panels, alongside the corresponding Grid score tiles, which reflect varying levels of Gaussian noise intensity. As noise intensity (NoI) increases, the grid scores generally decrease for both the RNN [Figs.~\hyperref[fig9]{\ref{fig9}(a)}] and leaky RNN [Figs.~\hyperref[fig9]{\ref{fig9}(b)} models (see also Fig.~\hyperref[fig8]{\ref{fig8}}). However, the leaky RNN consistently outperforms the RNN, maintaining more reliable grid patterns, particularly in the presence of Gaussian noise, where its grid structure remains robust despite the perturbations. Additionally, Figs.~\hyperref[fig9]{\ref{fig9}(c),(d)}  show similar trends under OU noise, with the leaky RNN  [Figs.~\hyperref[fig9]{\ref{fig9}(d)} again exhibiting more stable and well-defined grid patterns compared to the RNN  [Figs.~\hyperref[fig9]{\ref{fig9}(d)} under the same noise conditions. Notably, the grid scores for the leaky RNN consistently remain higher than those of the RNN, further reinforcing the conclusion that the  leaky RNN  is more resilient in preserving grid-like firing patterns in the presence of noise. These findings are further corroborated by the SAC analysis, presented in the bottom panels of Figs.~\hyperref[fig9]{\ref{fig9}(b,d)}.

\subsubsection{Two-dimensional attractor dynamics underlies path integration in the vanilla RNN and leaky RNN}

\begin{figure}[htp]	
	\includegraphics[width=\linewidth]{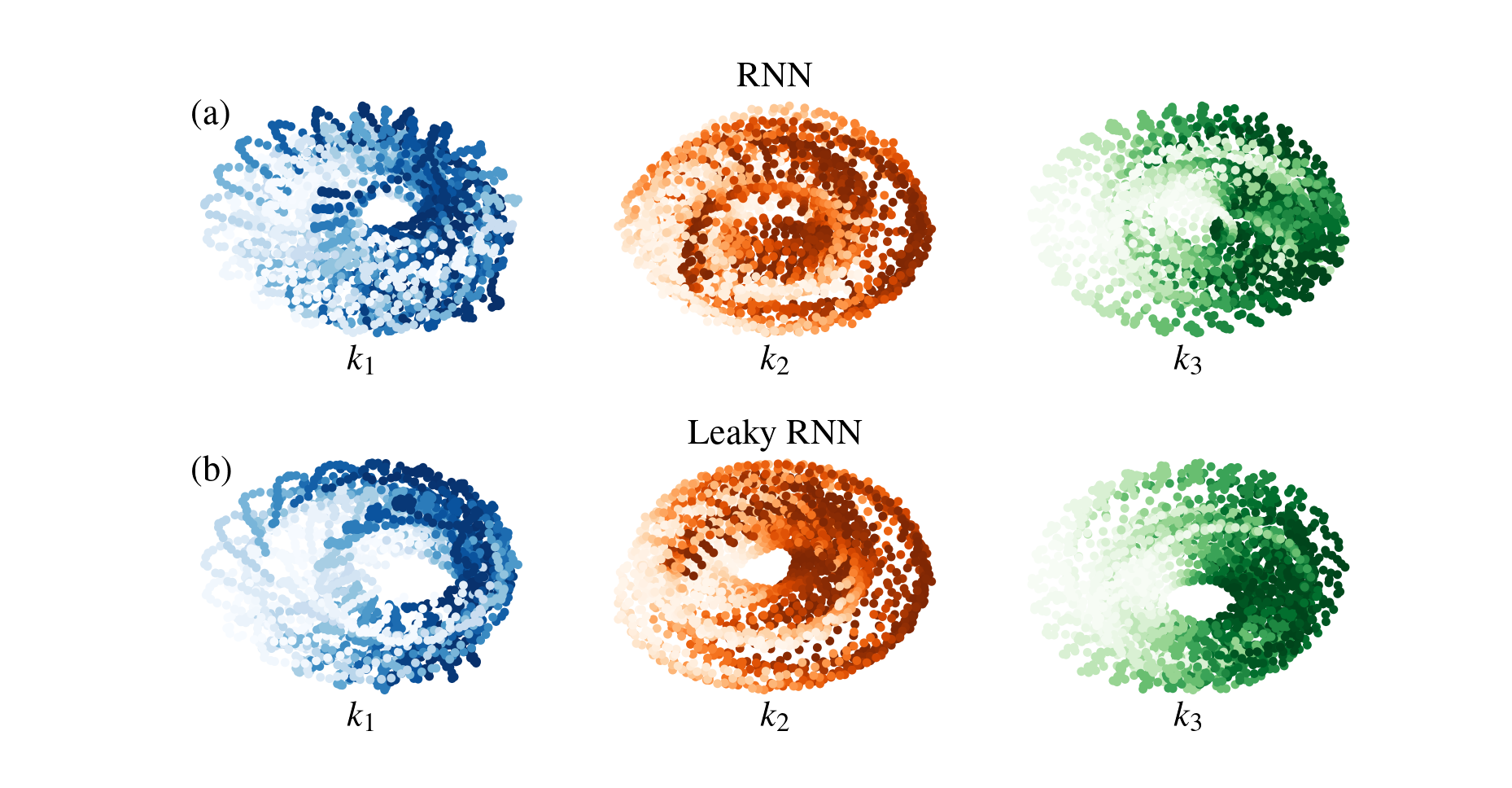}	  
	\caption{Projections of the toroidal manifold onto the three principal axes, $k_{1}$ (blue rings), $k_{2}$ (origin rings) and $k_3$ (green rings), reveal three distinct rings in both the RNN (top plots) and leaky RNN(bottom plots), corresponding to positions along the $0^\circ$,  $60^\circ$ and  $120^\circ$ vectors in real space. The colors represent the decoded position within the 2D environment.}
	\label{fig10}
\end{figure} 
Various previous studies have employed a 2D attractor manifold of stable activity patterns, structured as a torus, to maintain a memory trace of position while the animal remains stationary, without any velocity or other sensory inputs \cite{sorscher2023unified, GAORecurrent, conklin2005controlled, fuhs2006spin, ocko2018emergence}. This approach allows the system to preserve a stable representation of the animal's position in space, even in the absence of external movement cues. Building on this idea, we apply the Fourier power spectral density (PSD) to conduct an in-depth state-space analysis of both the RNN and leaky RNN models. The PSD of the neural maps reveals peaks that occur in a hexagonal arrangement along three principal axes, $k_{1}$,$k_{2}$ and $k_3$. Along these axes, the Fourier phases of each neuron, $\phi_1$, $\phi_2$ and $\phi_3$, exhibit a linear distribution. The PSD analysis allows us to examine the frequency components of the network's activity, revealing that the neurons spontaneously organize into stable attractor patterns, as clearly illustrated in Fig.~\hyperref[fig10]{\ref{fig10}}. 

Figure.~\hyperref[fig10]{\ref{fig10}} displays the torus attractors obtained from both the RNN and leaky RNN models. Along the first axis, $k_1$, at $0^\circ$, two stable tori are observed in both models. However, projections of the toroidal manifold along the $k_2$ (at $60^\circ$) and $k_3$ (at $120^\circ$) axes reveal that the stable torus with a distinct hole in the center is only present in the leaky RNN model. This feature is absent in the attractors generated by the RNN. These results further confirm that the leaky RNN models are more capable than RNN models in producing stable and regular grid-like firing patterns.

\begin{figure}[htp]
    \centering
    \includegraphics[width=\linewidth]{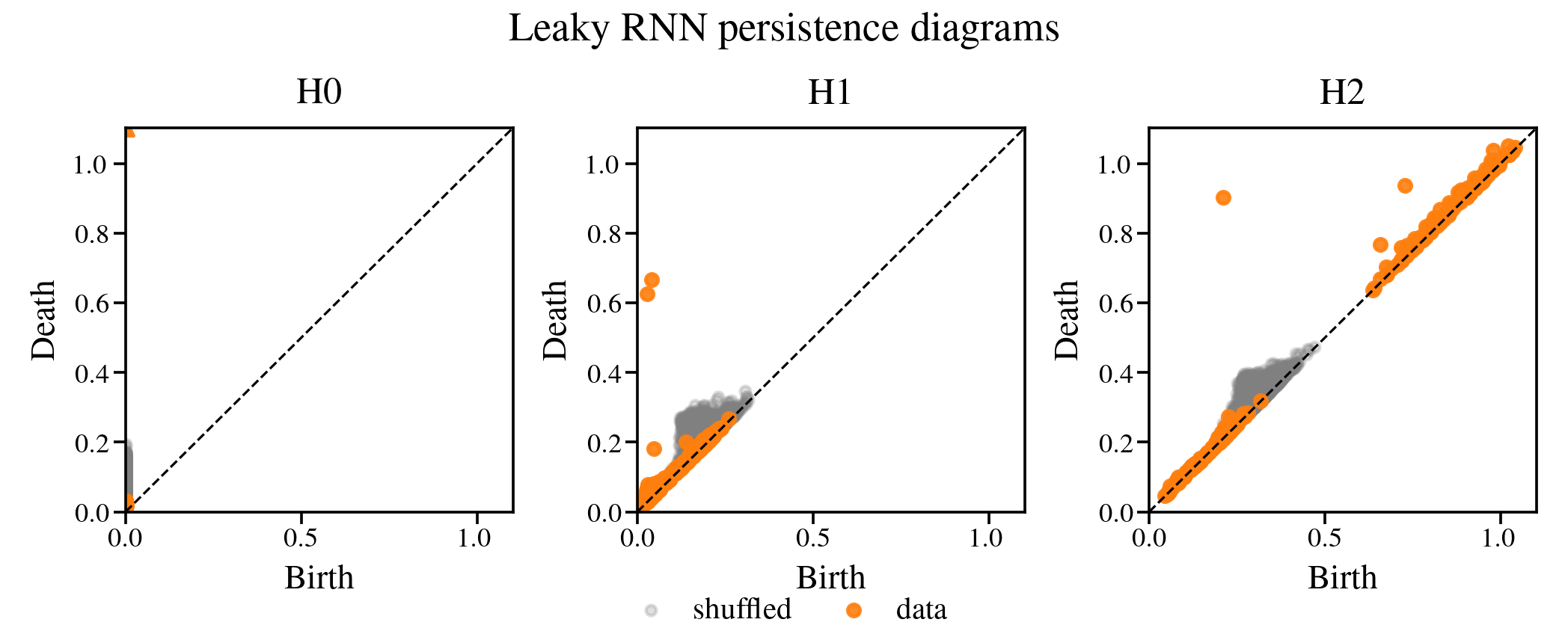}
    \caption{Persistence diagrams of the leaky RNN. Gray and orange points denote persistent homology obtained from shuffled control data.}
    \label{fig11}
\end{figure}
The main findings observed in Fig.~\hyperref[fig10]{\ref{fig10}} are also captured by persistent homology analysis. Figure~\ref{fig11} shows the persistence diagrams of neural activity generated by the leaky RNN during training. A long-lived feature appears in $H_0$, indicating that the neural activity manifold remains globally connected, whereas the remaining $H_0$ features exhibit short lifetimes, suggesting rapid merging of local connected components at small scales. In contrast, several long-lived features emerge in the $H_1$ and $H_2$ dimensions and are clearly separated from both the shuffled background and the diagonal, indicating that the leaky RNN forms a population activity manifold characterized by stable loop structures and higher-dimensional cavities.

\begin{figure}[htbp]
    \centering
    \includegraphics[width=\linewidth]{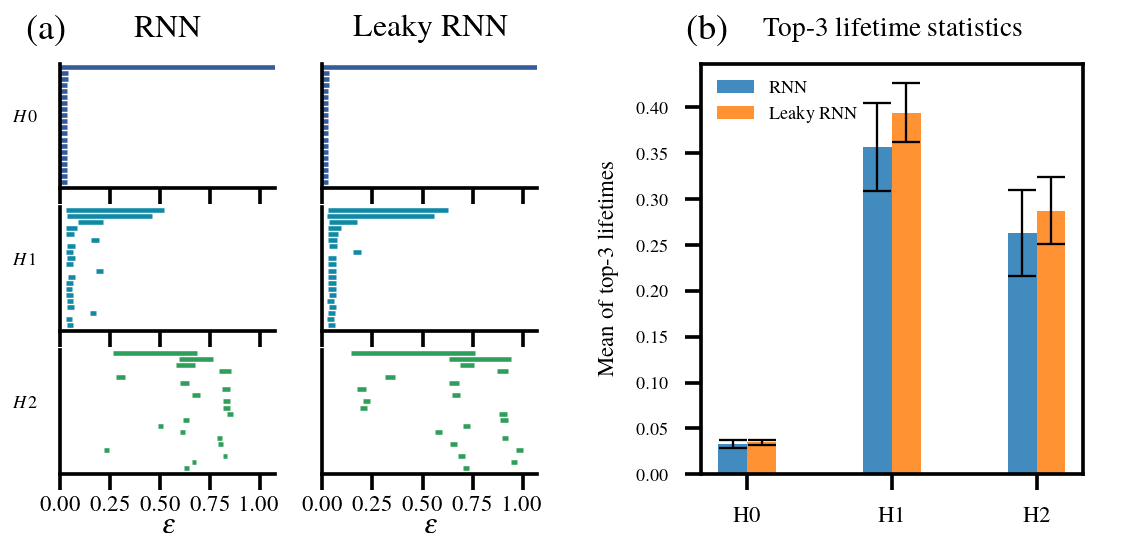}
    \caption{Comparison of Betti barcodes of neural activity manifolds in the RNN and Leaky RNN models.}
    \label{fig12}
\end{figure}

The neural activity manifolds of the RNN and Leaky RNN were further compared using Betti barcodes [Fig.~\hyperref[fig12]{\ref{fig12}(a)}]. As the connectivity scale increased, both models exhibited a persistent bar in the $H_0$ dimension, indicating that their neural activity manifolds remained globally connected. Notably, the Leaky RNN displayed clearer and longer-lived bars in the $H_1$ and $H_2$ dimensions, suggesting more stable loop structures and higher-dimensional cavities within its population activity. To quantitatively assess the stability of these topological features, the average lifetimes of the top three finite-lived features in $H_0$, $H_1$, and $H_2$ were computed across multiple runs [Fig.~\hyperref[fig12]{\ref{fig12}(b)}]. The Leaky RNN exhibited longer lifetimes in both $H_1$ and $H_2$, indicating enhanced persistence of nontrivial topological structures. Together, these results suggest that while both models maintain global connectivity, the Leaky RNN supports more stable loop organization and higher-dimensional cavities in the neural activity manifold.

\subsubsection{The low pass filtering effect of the leak term enhances spatial representations}
The discrete-time implementation of a simple RC low-pass filter, $ y^{t} = (1-\alpha)y^{t-1} + \alpha x^t$ represents the simplest form of exponential smoothing, also known as the exponentially weighted moving average \cite{momeni2018introduction}. In this equation, $\alpha$ is the smoothing factor within the range $0 \le \alpha \le 1$.  When simple exponential smoothing is applied, the output sequence is simply the weighted average of the current observation.Values of $\alpha$ close to 1  result in less smoothing, placing greater weight on recent changes in the data. Conversely, values of $\alpha$ closer to $0$ yield a stronger smoothing effect, making the system less responsive to recent changes. In other words, as $\alpha$ decreases, the output samples respond more slowly to changes in the input, as the system exhibits greater inertia for dynamical stability, ensuring dynamical stability\cite{momeni2018introduction,rongala2021non}. 

The leaky RNN (given in Eq.~\hyperref[leakRNN]{\ref{leakRNN}}) reflects the effect of the RC component integrated into the neural units, with the dynamic leak acting as a low-pass filter on the output activity. Recurrent circuitry is widely distributed throughout the brain, and such circuits can be prone to stability issues. More broadly, instability can render internal representations of information unreliable. In this work, when the networks were driven by velocity inputs with frequency $\omega_t$ to simulate random changes in head direction or stochastic noise for sensory perception, the recurrent network activity tended to generate high-frequency of self-amplifying components, sometimes evident as distinct transients, which were not present in the input data. The addition of a leak term, based on continuous-time attractor models of firing rates, consistently mitigated the instability arising from random changes in head direction or stochastic sensory noise. Overall, the dynamic leak in RNNs serves the beneficial function of protecting the recurrent neuronal circuitry from noise, preventing instability, and improving the accuracy of path integration in recurrent neural networks.

 \subsubsection{A mechanistic interpretation of optimized leak constants linking internal Dynamics and environmental Structure}
 The use of RNNs without a leak term represents a simplified formulation that may limit both optimal performance and mechanistic insight into the underlying dynamical processes \cite{quax2020adaptive,tani2016exploring,regazzoni2024learning}. Previous studies have shown that three main approaches have been proposed to determine the rate constants governing network dynamics in order to enhance the expressive capability of RNNs.The first approach optimizes networks using a single rate constant within a discrete-time approximation obtained through numerical integration\cite{pearlmutter1995gradient}. The second approach learns the leak constants directly by applying the backpropagation-through-time algorithm\cite{quax2020adaptive}. The third approach seeks to optimize leak constants in a manner that yields rate constants consistent with biologically plausible time scales \cite{tani2014self,quax2020adaptive}. Inspired by the third approach, we propose that optimizing leak constants may be mechanistically linked to temporal processes underlying internal neural dynamics, such as synaptic transmission and synaptic coupling. These processes interact strongly with external inputs, suggesting that leak optimization enables recurrent networks to align intrinsic dynamics with input-driven temporal structure.

\section{discussion}

The leaky RNN framework points toward an optimal leak strength that defines intrinsic time scales, which are associated with the emergence of periodic spatial activity of increasing complexity. To establish a relationship between the leak term and path integration accuracy, it is necessary to specify the temporal and spatial parameters of the leaky RNN used in this work. This enables the extraction of characteristic time scales for grid development and facilitates comparison with previous theoretical and experimental findings\cite{burak2009accurate,stella2015self}. Within a biologically plausible parameter regime, continuous attractor models of path integration in the dorsolateral medial entorhinal cortex suggest that such networks can achieve accurate path integration over time scales of approximately $1$–$10$ minutes and spatial ranges of $10$–$100$ meters\cite{burak2009accurate}. Moreover, experimental study in bats navigating three-dimensional environments has revealed that the time scale of grid formation depends on both environmental structure and size. For example, in a small cubic enclosure, time scales of grid formation is nearly 3 hours,  such as triplets of firing fields forming approximately equilateral triangles emerge after $10$–$12$ hours of continuous flight,followed by the appearance of hexagonal patterns after $15$–$18$ hours. Eventually, different grid units align to a common orientation after approximately $30$–$35$ hours\cite{stella2015self}. These findings suggest that  time scales of grid formation are not fixed but can vary within the same neural model depending on environmental and spatial factors. Consequently, the development of grid-like firing patterns relative to grid spacing,  is governed by an optimal time scale (optimal leak rate $\alpha$) in the leaky RNN.

In this study, we investigate the combined effects of leak-induced time scales and ReLU activation functions in recurrent neural networks to facilitate the emergence of grid-like firing patterns characteristic of spatial navigation systems. Specifically, we demonstrate that incorporating a leak term, parameterized by $\alpha$, enhances the network’s ability to generate stable and well-structured grid representations. The introduced temporal dynamics enable more effective regulation of information flow across time steps, thereby improving the temporal stability of neural activity patterns.The ReLU activation function, owing to its strong nonlinearity and training efficiency, further promotes the formation of clear hexagonal firing patterns, a hallmark of grid-cell activity. Our results show that leaky RNNs significantly improve the emergence of highly regular hexagonal grid structures. These findings are supported by both Grid Score analyses\cite{gardner2022toroidal} and spatial autocorrelation measurements\cite{legendre1993spatial,anselin2003introduction,griffith2025spatial,oyana2020spatial}, which confirm increased symmetry and regularity in the firing patterns produced by leaky RNNs. Notably, the grid-like activity encoded by leaky RNNs  exhibits clear and evenly spaced peaks, indicative of a stable and robust grid organization. 

Furthermore, we compare the position estimates derived from the leaky RNN with the true trajectories to assess its accuracy. Using the mean squared error\cite{hodson2022root} as a metric, we find that the trained leaky RNN provides more accurate position predictions, which is critical for the reliable generation of grid-cell-like firing patterns. The improved accuracy in position estimation directly correlates with the ability of the leaky RNN to maintain grid-like firing structures, even in the presence of noise, making it a more robust model for spatial navigation tasks.The leaky RNN consistently outperforms the standard RNN, particularly in its ability to produce more stable and well-defined grid patterns under identical noise conditions. This enhanced performance is especially evident in the comparison of grid patterns between the two models, where the leaky RNN exhibits more precise and stable hexagonal arrangements, even when noise is introduced into the system.

Finally, we examine the torus attractors generated by both the RNN and leaky RNN models, which visually represent the dynamic behavior of the systems. The analysis of these attractors shows that the leaky RNN, with its distinct central hole in the attractor plot, exhibits greater stability and regularity in generating grid-like firing patterns. This feature suggests that the leaky RNN's architecture supports better organization and stability of neural activity, thereby enhancing its ability to produce reliable grid-cell-like firing patterns compared to the standard RNN. However, the roles of excitatory-inhibitory balance\cite{zhang2025perturbation,weber2018learning, liu2025formation} and synaptic plasticity\cite{zhou2022short,widloski2014model,mehta2015synaptic,shan2025short}, which may contribute to the emergence of regular grid-like firing patterns, have not been extensively explored in this study.

The coordinated actions of excitation and inhibition within neural circuits have primarily been studied through the perspectives of synaptic weight tuning and synaptic plasticity \cite{zhang2025perturbation, liu2025formation, zhou2022short, widloski2014model, mehta2015synaptic, zhu2024chaos,shan2025short}. Some research has explored the neural mechanisms underlying path-integration-based navigation in the context of synaptic weights and plasticity \cite{weber2018learning, christensen2021perineuronal, wang2023intrinsic}. Notably, both excitatory and inhibitory synaptic plasticity play critical roles in the spatial selectivities and invariances seen in grid cells, place cells, and head-direction cells. The interaction of excitatory and inhibitory plasticity can give rise to localized, grid-like, or invariant activity patterns \cite{weber2018learning}. Moreover, recent studies suggest that any circularly symmetric, inhibition-dominated local interaction profile of sufficient strength can produce the characteristic hexagonal grid patterns \cite{khona2021spontaneous}. Regardless of the specific form of network interactions, the presence of both excitation and inhibition is essential for stabilizing firing patterns \cite{dong2024grid,gong2026role}.

\begin{figure}[htp]	
	\includegraphics[width=\linewidth]{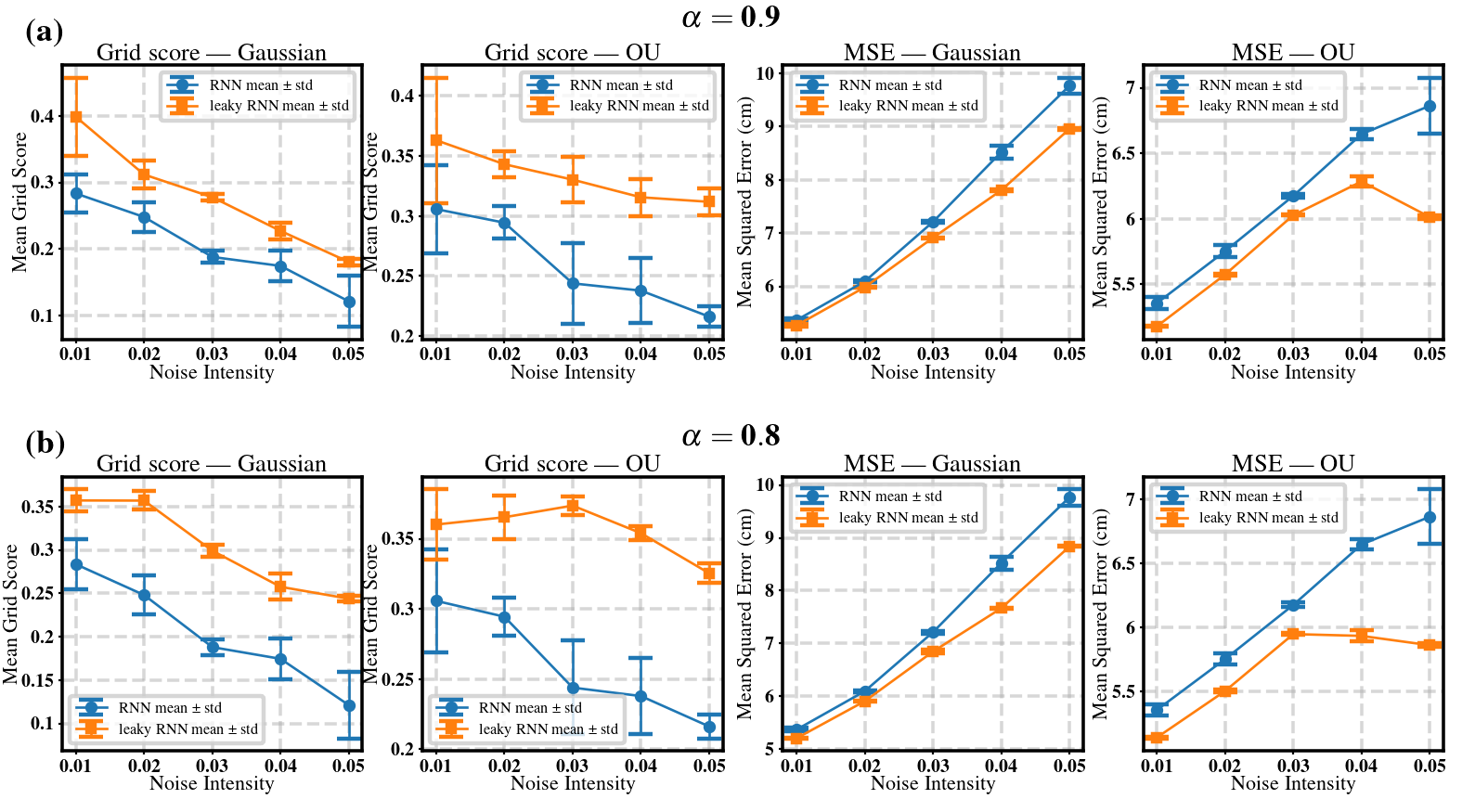}	  
	\caption{The grid scores and mean squared errors (MSE) as functions of noise intensity for $\alpha = 0.90$ and $\alpha = 0.80$  are evaluated under both Gaussian white noise and OU noise.}
	\label{fig13}
\end{figure}

The role of chaotic behavior in recurrent units and its contribution to the emergence of grid-cell-like firing patterns remains subtle and elusive. Recurrent neural networks are designed with tuned recurrent connections, operating in a dynamic regime that supports the coexistence of chaotic and locally stable trajectories. By selectively establishing stable dynamic attractors, this approach leverages both the rich dynamics of chaotic networks and the reproducibility of deterministic ones \cite{laje2013robust}. Networks of chaotic cells have been shown to outperform networks of non-chaotic cells in tasks involving complete synchronization, especially when short-term plasticity is considered \cite{shan2025short}. The role of chaotic units in the emergence and maintenance of grid-like firing patterns in recurrent neural networks will be the focus of future work.

The proposed leaky RNN model still relies on several simplifying assumptions. First, place cells are treated as supervisory signals for path integration, which may not fully capture the biological mechanisms underlying spatial representation. Second, the model exhibits only a single spatial scale, whereas biological grid cells are organized across multiple spatial scales. In addition, the current framework primarily depends on self-motion input and lacks an explicit sensory correction mechanism; incorporating such feedback in future work could improve both robustness and biological plausibility. Finally, the leak parameter $\alpha$ remains sensitive to changes in spatial scale, suggesting that future studies should explore adaptive mechanisms that better align firing-pattern scale with environmental scale.

This work opens several promising avenues for further investigation. Our findings highlight how combining leaky recurrent units with ReLU activation functions within recurrent neural networks enhances the emergence of grid-like firing patterns, a hallmark of spatial navigation systems.

\begin{acknowledgements}
The authors thank all members of the HQU–BUAS–JSU–NBU–BNU–JXU–USST Joint Group for their valuable discussions on neuronal dynamics. This work was supported by the National Natural Science Foundation of China (Grant No. 12165016) and the Scientific Research Startup Fund of Jiangsu University (Grant No. 5501190017). M. Zheng acknowledges support from the Jiangsu Specially Appointed Professor Program.
\end{acknowledgements}


\section*{Data availability}
The data will be uploaded to GitHub after acceptance.

 \section*{Conflict of interest}
The authors declare that they have no conflict of interest.

\appendix

\section{Grid scores and coding errors across different levels of noise intensity}\label{secA1}

The grid scores and coding errors of the leaky RNN were evaluated across a range of noise intensities under both Gaussian and OU noise, alongside results from standard RNNs with $\alpha = 0.9$ (Figs.\hyperref[fig13]{\ref{fig13}}(a)) and $\alpha = 0.8$ (Figs.\hyperref[fig13]{\ref{fig13}}(b)). As shown in these figures, the leaky RNN consistently achieves higher grid scores than the RNN across all levels of noise intensity and under both noise types.

\section{Numerical simulation}\label{secB1}
To ensure reproducibility, random seeds were fixed across all stochastic processes. To isolate the effect of model architecture, the weights of the standard RNN were initialized at the beginning of each trial, and the identical initial parameters were subsequently assigned to the leaky RNN. All remaining experimental factors, including optimizer settings, trajectory generation procedures, and training protocols, were strictly matched between the two models, ensuring identical optimization conditions and comparable starting states. For each parameter configuration, five independent trials were performed and the results were averaged. Under these controlled conditions, observed reductions in mean squared errors and improvements in firing pattern regularity can be causally attributed to the structural modification introduced by the leaky RNN.

\bibliographystyle{unsrt}

\bibliography{AI_gridcell}   

@article{hafting2005microstructure,
	title={Microstructure of a spatial map in the entorhinal cortex},
	author={Hafting, Torkel and Fyhn, Marianne and Molden, Sturla and Moser, May-Britt and Moser, Edvard I},
	journal={Nature},
	volume={436},
	number={7052},
	pages={801--806},
	year={2005},
	publisher={Nature Publishing Group UK London}
}

@article{rebecca2025spatial,
	title={Spatial periodicity in grid cell firing is explained by a neural sequence code of 2-D trajectories},
	author={Rebecca, RG and Ascoli, Giorgio A and Sutton, Nate M and Dannenberg, Holger},
	journal={Elife},
	volume={13},
	pages={RP96627},
	year={2025},
	publisher={eLife Sciences Publications Limited}
}

@article{moser2008place,
	title={Place cells, grid cells, and the brain's spatial representation system},
	author={Moser, Edvard I and Kropff, Emilio and Moser, May-Britt},
	journal={Annu. Rev. Neurosci.},
	volume={31},
	number={1},
	pages={69--89},
	year={2008},
	publisher={Annual Reviews}
}

@article{dong2024grid,
	title={Grid cells in cognition: mechanisms and function},
	author={Dong, Ling L and Fiete, Ila R},
	journal={Annual Review of Neuroscience},
	volume={47},
	year={2024},
	publisher={Annual Reviews}
}

@article{rowland2016ten,
	title={Ten years of grid cells},
	author={Rowland, David C and Roudi, Yasser and Moser, May-Britt and Moser, Edvard I},
	journal={Annual review of neuroscience},
	volume={39},
	number={1},
	pages={19--40},
	year={2016},
	publisher={Annual Reviews}
}

@article{birch2007rectangular,
	title={Rectangular and hexagonal grids used for observation, experiment and simulation in ecology},
	author={Birch, Colin PD and Oom, Sander P and Beecham, Jonathan A},
	journal={Ecological modelling},
	volume={206},
	number={3-4},
	pages={347--359},
	year={2007},
	publisher={Elsevier}
}

@article{moser2014grid,
	title={Grid cells and cortical representation},
	author={Moser, Edvard I and Roudi, Yasser and Witter, Menno P and Kentros, Clifford and Bonhoeffer, Tobias and Moser, May-Britt},
	journal={Nature Reviews Neuroscience},
	volume={15},
	number={7},
	pages={466--481},
	year={2014},
	publisher={Nature Publishing Group UK London}
}

@article{doeller2010evidence,
	title={Evidence for grid cells in a human memory network},
	author={Doeller, Christian F and Barry, Caswell and Burgess, Neil},
	journal={Nature},
	volume={463},
	number={7281},
	pages={657--661},
	year={2010},
	publisher={Nature Publishing Group UK London}
}

@article{morris2023chicken,
	title={The chicken and egg problem of grid cells and place cells},
	author={Morris, Genela and Derdikman, Dori},
	journal={Trends in Cognitive Sciences},
	volume={27},
	number={2},
	pages={125--138},
	year={2023},
	publisher={Elsevier}
}

@article{bush2014grid,
	title={What do grid cells contribute to place cell firing?},
	author={Bush, Daniel and Barry, Caswell and Burgess, Neil},
	journal={Trends in neurosciences},
	volume={37},
	number={3},
	pages={136--145},
	year={2014},
	publisher={Elsevier}
}

@article{o1979review,
	title={A review of the hippocampal place cells},
	author={O'Keefe, John},
	journal={Progress in neurobiology},
	volume={13},
	number={4},
	pages={419--439},
	year={1979},
	publisher={Elsevier}
}

@article{kropff2015speed,
	title={Speed cells in the medial entorhinal cortex},
	author={Kropff, Emilio and Carmichael, James E and Moser, May-Britt and Moser, Edvard I},
	journal={Nature},
	volume={523},
	number={7561},
	pages={419--424},
	year={2015},
	publisher={Nature Publishing Group UK London}
}

@article{taube2007head,
	title={The head direction signal: origins and sensory-motor integration},
	author={Taube, Jeffrey S},
	journal={Annu. Rev. Neurosci.},
	volume={30},
	number={1},
	pages={181--207},
	year={2007},
	publisher={Annual Reviews}
}

@article{poulter2018neurobiology,
	title={The neurobiology of mammalian navigation},
	author={Poulter, Steven and Hartley, Tom and Lever, Colin},
	journal={Current Biology},
	volume={28},
	number={17},
	pages={R1023--R1042},
	year={2018},
	publisher={Elsevier}
}

@article{solstad2008representation,
	title={Representation of geometric borders in the entorhinal cortex},
	author={Solstad, Trygve and Boccara, Charlotte N and Kropff, Emilio and Moser, May-Britt and Moser, Edvard I},
	journal={Science},
	volume={322},
	number={5909},
	pages={1865--1868},
	year={2008},
	publisher={American Association for the Advancement of Science}
}

@article{sorscher2023unified,
	title={A unified theory for the computational and mechanistic origins of grid cells},
	author={Sorscher, Ben and Mel, Gabriel C and Ocko, Samuel A and Giocomo, Lisa M and Ganguli, Surya},
	journal={Neuron},
	volume={111},
	number={1},
	pages={121--137},
	year={2023},
	publisher={Elsevier}
}

@article{giocomo2011computational,
	title={Computational models of grid cells},
	author={Giocomo, Lisa M and Moser, May-Britt and Moser, Edvard I},
	journal={Neuron},
	volume={71},
	number={4},
	pages={589--603},
	year={2011},
	publisher={Elsevier}
}

@article{o2005dual,
	title={Dual phase and rate coding in hippocampal place cells: theoretical significance and relationship to entorhinal grid cells},
	author={O'keefe, John and Burgess, Neil},
	journal={Hippocampus},
	volume={15},
	number={7},
	pages={853--866},
	year={2005},
	publisher={Wiley Online Library}
}

@article{mcnaughton2006path,
	title={Path integration and the neural basis of the'cognitive map'},
	author={McNaughton, Bruce L and Battaglia, Francesco P and Jensen, Ole and Moser, Edvard I and Moser, May-Britt},
	journal={Nature Reviews Neuroscience},
	volume={7},
	number={8},
	pages={663--678},
	year={2006},
	publisher={Nature Publishing Group UK London}
}

@article{GAORecurrent,
	title = {Recurrent spiking neural networks as models of the entorhinal–hippocampal system for path integration: Grid cells and beyond},
	journal = {Neurocomputing},
	volume = {651},
	pages = {130814},
	year = {2025},
	author = {Ruilan Gao and Changjian Jiang and Yu Zhang},
}

@article{campbell2018principles,
	title={Principles governing the integration of landmark and self-motion cues in entorhinal cortical codes for navigation},
	author={Campbell, Malcolm G and Ocko, Samuel A and Mallory, Caitlin S and Low, Isabel IC and Ganguli, Surya and Giocomo, Lisa M},
	journal={Nature neuroscience},
	volume={21},
	number={8},
	pages={1096--1106},
	year={2018},
	publisher={Nature Publishing Group US New York}
}

@article{widloski2014model,
	title={A model of grid cell development through spatial exploration and spike time-dependent plasticity},
	author={Widloski, John and Fiete, Ila R},
	journal={Neuron},
	volume={83},
	number={2},
	pages={481--495},
	year={2014},
	publisher={Elsevier}
}

@article{ocko2018emergent,
	title={Emergent elasticity in the neural code for space},
	author={Ocko, Samuel A and Hardcastle, Kiah and Giocomo, Lisa M and Ganguli, Surya},
	journal={Proceedings of the National Academy of Sciences},
	volume={115},
	number={50},
	pages={E11798--E11806},
	year={2018},
	publisher={National Academy of Sciences}}

@article{banino2018vector,
title={Vector-based navigation using grid-like representations in artificial agents},
author={Banino, Andrea and Barry, Caswell and Uria, Benigno and Blundell, Charles and Lillicrap, Timothy and Mirowski, Piotr and Pritzel, Alexander and Chadwick, Martin J and Degris, Thomas and Modayil, Joseph and others},
journal={Nature},
volume={557},
number={7705},
pages={429--433},
year={2018},
publisher={Nature Publishing Group UK London}
}

@article{kanitscheider2017training,
	title={Training recurrent networks to generate hypotheses about how the brain solves hard navigation problems},
	author={Kanitscheider, Ingmar and Fiete, Ila},
	journal={Advances in Neural Information Processing Systems},
	volume={30},
	year={2017}
}

@article{sorscher2019unified,
	title={A unified theory for the origin of grid cells through the lens of pattern formation},
	author={Sorscher, Ben and Mel, Gabriel and Ganguli, Surya and Ocko, Samuel},
	journal={Advances in neural information processing systems},
	volume={32},
	year={2019}
}

@article{fernandez2024grid,
	title={The grid-cell normative model: Unifying ‘principles’},
	author={Fernandez-Leon, Jose A and Sarramone, Luca},
	journal={BioSystems},
	volume={235},
	pages={105091},
	year={2024},
	publisher={Elsevier}
}

@inproceedings{dey2017gate,
	title={Gate-variants of gated recurrent unit (GRU) neural networks},
	author={Dey, Rahul and Salem, Fathi M},
	booktitle={2017 IEEE 60th international midwest symposium on circuits and systems (MWSCAS)},
	pages={1597--1600},
	year={2017},
	organization={IEEE}
}

@article{chung2014empirical,
	title={Empirical evaluation of gated recurrent neural networks on sequence modeling},
	author={Chung, Junyoung and Gulcehre, Caglar and Cho, KyungHyun and Bengio, Yoshua},
	journal={arXiv preprint arXiv:1412.3555},
	year={2014}
}

@incollection{salem2021gated,
	title={Gated RNN: the gated recurrent unit (GRU) RNN},
	author={Salem, Fathi M},
	booktitle={Recurrent neural networks: from simple to gated architectures},
	pages={85--100},
	year={2021},
	publisher={Springer}
}

@article{cho2014learning,
	title={Learning phrase representations using RNN encoder-decoder for statistical machine translation},
	author={Cho, Kyunghyun and Van Merri{\"e}nboer, Bart and Gulcehre, Caglar and Bahdanau, Dzmitry and Bougares, Fethi and Schwenk, Holger and Bengio, Yoshua},
	journal={arXiv preprint arXiv:1406.1078},
	year={2014}
}

@article{chaplot2018active,
	title={Active neural localization},
	author={Chaplot, Devendra Singh and Parisotto, Emilio and Salakhutdinov, Ruslan},
	journal={arXiv preprint arXiv:1801.08214},
	year={2018}
}

@inproceedings{glorot2010understanding,
	title={Understanding the difficulty of training deep feedforward neural networks},
	author={Glorot, Xavier and Bengio, Yoshua},
	booktitle={Proceedings of the thirteenth international conference on artificial intelligence and statistics},
	pages={249--256},
	year={2010},
	organization={JMLR Workshop and Conference Proceedings}
}

@article{connor2024correlations,
	title={Correlations of cross-entropy loss in machine learning},
	author={Connor, Richard and Dearle, Alan and Claydon, Ben and Vadicamo, Lucia},
	journal={Entropy},
	volume={26},
	number={6},
	pages={491},
	year={2024},
	publisher={MDPI}
}

@article{conklin2005controlled,
	title={A controlled attractor network model of path integration in the rat},
	author={Conklin, John and Eliasmith, Chris},
	journal={Journal of computational neuroscience},
	volume={18},
	number={2},
	pages={183--203},
	year={2005},
	publisher={Springer}
}

@article{ocko2018emergence,
	title={The emergence of multiple retinal cell types through efficient coding of natural movies},
	author={Ocko, Samuel and Lindsey, Jack and Ganguli, Surya and Deny, Stephane},
	journal={Advances in Neural Information Processing Systems},
	volume={31},
	year={2018}
}

@article{gardner2022toroidal,
	title={Toroidal topology of population activity in grid cells},
	author={Gardner, Richard J and Hermansen, Erik and Pachitariu, Marius and Burak, Yoram and Baas, Nils A and Dunn, Benjamin A and Moser, May-Britt and Moser, Edvard I},
	journal={Nature},
	volume={602},
	number={7895},
	pages={123--128},
	year={2022},
	publisher={Nature Publishing Group UK London}
}

@article{legendre1993spatial,
	title={Spatial autocorrelation: trouble or new paradigm?},
	author={Legendre, Pierre},
	journal={Ecology},
	volume={74},
	number={6},
	pages={1659--1673},
	year={1993},
	publisher={Wiley Online Library}
}

@article{anselin2003introduction,
	title={An introduction to spatial autocorrelation analysis with GeoDa},
	author={Anselin, Luc},
	journal={Spatial Analysis Laboratory, University of Illinois, Champagne-Urbana, Illinois},
	year={2003}
}

@book{griffith2025spatial,
	title={Spatial Autocorrelation: A Fundamental Property of Geospatial Phenomena},
	author={Griffith, Daniel and Li, Bin},
	year={2025},
	publisher={Elsevier}
}

@book{oyana2020spatial,
	title={Spatial analysis with R: statistics, visualization, and computational methods},
	author={Oyana, Tonny J},
	year={2020},
	publisher={CRC press}
}

@article{hodson2022root,
	title={Root mean square error (RMSE) or mean absolute error (MAE): When to use them or not},
	author={Hodson, Timothy O},
	journal={Geoscientific Model Development Discussions},
	volume={2022},
	pages={1--10},
	year={2022},
	publisher={G{\"o}ttingen, Germany}
}

@article{shan2025short,
	title={Short-term plasticity promotes synchronization of coupled chaotic oscillators in excitatory--inhibitory networks},
	author={Shan, Kun and Tian, Changhai and Zheng, Zhigang and Zheng, Muhua and Xu, Kesheng},
	journal={Chaos: An Interdisciplinary Journal of Nonlinear Science},
	volume={35},
	number={7},
	year={2025},
	publisher={AIP Publishing}
}

@article{zhang2025perturbation,
	title={Perturbation-driven of stability and phase transition in balanced rate-based networks},
	author={Zhang, Huiwen and Zhang, Yan and Wu, Jiao and Zheng, Muhua and Xu, kesheng},
	journal={The European Physical Journal Special Topics},
	pages={1--15},
	year={2025},
	publisher={Springer}
}

@article{liu2025formation,
	title={Formation and maintenance of neuronal collective dynamics through local perturbation and intrinsic node dynamics},
	author={Liu, Runzhou and Zhang, Yan and Wang, Jijun and Zheng, Muhua and Xu, Kesheng},
	journal={Chaos, Solitons \& Fractals},
	volume={199},
	pages={116648},
	year={2025},
	publisher={Elsevier}
}

@article{weber2018learning,
	title={Learning place cells, grid cells and invariances with excitatory and inhibitory plasticity},
	author={Weber, Simon Nikolaus and Sprekeler, Henning},
	journal={Elife},
	volume={7},
	pages={e34560},
	year={2018},
	publisher={eLife Sciences Publications, Ltd}
}

@article{zhou2022short,
	title={Short-term plasticity as a mechanism to regulate and retain multistability},
	author={Zhou, Xinjia and Tian, Changhai and Zhang, Xiyun and Zheng, Muhua and Xu, Kesheng},
	journal={Chaos, Solitons \& Fractals},
	volume={165},
	pages={112891},
	year={2022},
	publisher={Elsevier}
}

@article{mehta2015synaptic,
	title={From synaptic plasticity to spatial maps and sequence learning},
	author={Mehta, Mayank R},
	journal={Hippocampus},
	volume={25},
	number={6},
	pages={756--762},
	year={2015},
	publisher={Wiley Online Library}
}

@article{christensen2021perineuronal,
	title={Perineuronal nets stabilize the grid cell network},
	author={Christensen, Ane Charlotte and Lensj{\o}, Kristian Kinden and Lepper{\o}d, Mikkel Elle and Dragly, Svenn-Arne and Sutterud, Halvard and Blackstad, Jan Sigurd and Fyhn, Marianne and Hafting, Torkel},
	journal={Nature Communications},
	volume={12},
	number={1},
	pages={253},
	year={2021},
	publisher={Nature Publishing Group UK London}
}

@article{wang2023intrinsic,
	title={Intrinsic theta oscillation in the attractor network of grid cells},
	author={Wang, Ziqun and Wang, Tao and Yang, Fan and Liu, Feng and Wang, Wei},
	journal={Iscience},
	volume={26},
	number={4},
	year={2023},
	publisher={Elsevier}
}

@article{khona2021spontaneous,
	title={Spontaneous emergence of topologically robust grid cell modules: A multiscale instability theory},
	author={Khona, Mikail and Chandra, Sarthak and Fiete, Ila},
	journal={bioRxiv},
	year={2021},
	publisher={Cold Spring Harbor Laboratory}
}

@article{laje2013robust,
	title={Robust timing and motor patterns by taming chaos in recurrent neural networks},
	author={Laje, Rodrigo and Buonomano, Dean V},
	journal={Nature neuroscience},
	volume={16},
	number={7},
	pages={925--933},
	year={2013},
	publisher={Nature Publishing Group US New York}
}

@article{zhu2024chaos,
	title={Chaos shapes transient synchrony activities and switchings in the excitatory-inhibitory networks},
	author={Zhu, Gaobiao and Zhang, Yan and Wu, Jiao and Zheng, Muhua and Xu, Kesheng},
	journal={Nonlinear Dynamics},
	volume={112},
	number={9},
	pages={7555--7570},
	year={2024},
	publisher={Springer}
}

@article{banino_vectorbased_2018,
  title = {Vector-Based Navigation Using Grid-like Representations in Artificial Agents},
  author = {Banino, Andrea and Barry, Caswell and Uria, Benigno and Blundell, Charles and Lillicrap, Timothy and Mirowski, Piotr and Pritzel, Alexander and Chadwick, Martin J. and Degris, Thomas and Modayil, Joseph and Wayne, Greg and Soyer, Hubert and Viola, Fabio and Zhang, Brian and Goroshin, Ross and Rabinowitz, Neil and Pascanu, Razvan and Beattie, Charlie and Petersen, Stig and Sadik, Amir and Gaffney, Stephen and King, Helen and Kavukcuoglu, Koray and Hassabis, Demis and Hadsell, Raia and Kumaran, Dharshan},
  year = 2018,
  month = may,
  journal = {Nature},
  volume = {557},
  number = {7705},
  pages = {429--433},
  publisher = {Nature Publishing Group},
  issn = {1476-4687},
  urldate = {2026-03-03},
  copyright = {2018 Macmillan Publishers Ltd., part of Springer Nature},
  langid = {english}
}

@article{raudies2012modeling,
	title={Modeling boundary vector cell firing given optic flow as a cue},
	author={Raudies, Florian and Hasselmo, Michael E},
	journal={PLoS computational biology},
	volume={8},
	number={6},
	pages={e1002553},
	year={2012},
	publisher={Public Library of Science San Francisco, USA}
}

@article{sorscher_unified_2023,
  title = {A Unified Theory for the Computational and Mechanistic Origins of Grid Cells},
  author = {Sorscher, Ben and Mel, Gabriel C. and Ocko, Samuel A. and Giocomo, Lisa M. and Ganguli, Surya},
  year = 2023,
  month = jan,
  journal = {Neuron},
  volume = {111},
  number = {1},
  pages = {121-137.e13},
  publisher = {Elsevier},
  issn = {0896-6273},
  urldate = {2026-03-03},
  langid = {english},
  pmid = {36306779}
}

@article{JUN20201095,
title = {Disrupted Place Cell Remapping and Impaired Grid Cells in a Knockin Model of Alzheimer's Disease},
journal = {Neuron},
volume = {107},
number = {6},
pages = {1095-1112.e6},
year = {2020},
issn = {0896-6273},
doi = {https://doi.org/10.1016/j.neuron.2020.06.023},
author = {Heechul Jun and Allen Bramian and Shogo Soma and Takashi Saito and Takaomi C. Saido and Kei M. Igarashi},
}

@article{langston2010development,
	title={Development of the spatial representation system in the rat},
	author={Langston, Rosamund F and Ainge, James A and Couey, Jonathan J and Canto, Cathrin B and Bjerknes, Tale L and Witter, Menno P and Moser, Edvard I and Moser, May-Britt},
	journal={Science},
	volume={328},
	number={5985},
	pages={1576--1580},
	year={2010},
	publisher={American Association for the Advancement of Science}
}

@article{perez2016visual,
	title={Visual landmarks sharpen grid cell metric and confer context specificity to neurons of the medial entorhinal cortex},
	author={P{\'e}rez-Escobar, Jos{\'e} Antonio and Kornienko, Olga and Latuske, Patrick and Kohler, Laura and Allen, Kevin},
	journal={Elife},
	volume={5},
	pages={e16937},
	year={2016},
	publisher={eLife Sciences Publications, Ltd}
}

@article{burak2009accurate,
	title={Accurate path integration in continuous attractor network models of grid cells},
	author={Burak, Yoram and Fiete, Ila R},
	journal={PLoS computational biology},
	volume={5},
	number={2},
	pages={e1000291},
	year={2009},
	publisher={Public Library of Science San Francisco, USA}
}

@article{fuhs2006spin,
	title={A spin glass model of path integration in rat medial entorhinal cortex},
	author={Fuhs, Mark C and Touretzky, David S},
	journal={Journal of Neuroscience},
	volume={26},
	number={16},
	pages={4266--4276},
	year={2006},
	publisher={Society for Neuroscience}
}

@inproceedings{monfared2020transformation,
	title={Transformation of ReLU-based recurrent neural networks from discrete-time to continuous-time},
	author={Monfared, Zahra and Durstewitz, Daniel},
	booktitle={International Conference on Machine Learning},
	pages={6999--7009},
	year={2020},
	organization={PMLR}
}

@article{quax2020adaptive,
	title={Adaptive time scales in recurrent neural networks},
	author={Quax, Silvan C and D’asaro, Michele and Van Gerven, Marcel AJ},
	journal={Scientific reports},
	volume={10},
	number={1},
	pages={11360},
	year={2020},
	publisher={Nature Publishing Group UK London}
}

@inproceedings{linemergence,
	title={Emergence of Spatial Representation in an Actor-Critic Agent with Hippocampus-Inspired Sequence Generator},
	author={Lin, Xiao-Xiong and Yiu, Yuk-Hoi and Leibold, Christian},
	booktitle={The Fourteenth International Conference on Learning Representations}
}

@article{jaeger2007optimization,
	title={Optimization and applications of echo state networks with leaky-integrator neurons},
	author={Jaeger, Herbert and Luko{\v{s}}evi{\v{c}}ius, Mantas and Popovici, Dan and Siewert, Udo},
	journal={Neural networks},
	volume={20},
	number={3},
	pages={335--352},
	year={2007},
	publisher={Elsevier}
}

@article{jaeger2001echo,
	title={The “echo state” approach to analysing and training recurrent neural networks-with an erratum note},
	author={Jaeger, Herbert},
	journal={Bonn, Germany: German national research center for information technology gmd technical report},
	volume={148},
	number={34},
	pages={13},
	year={2001},
	publisher={Bonn}
}

@article{dordek2016extracting,
	title={Extracting grid cell characteristics from place cell inputs using non-negative principal component analysis},
	author={Dordek, Yedidyah and Soudry, Daniel and Meir, Ron and Derdikman, Dori},
	journal={Elife},
	volume={5},
	pages={e10094},
	year={2016},
	publisher={eLife Sciences Publications, Ltd}
}

@article{driscoll2024flexible,
	title={Flexible multitask computation in recurrent networks utilizes shared dynamical motifs},
	author={Driscoll, Laura N and Shenoy, Krishna and Sussillo, David},
	journal={Nature Neuroscience},
	volume={27},
	number={7},
	pages={1349--1363},
	year={2024},
	publisher={Nature Publishing Group US New York}
}

@book{momeni2018introduction,
	title={Introduction to statistical methods in pathology},
	author={Momeni, Amir and Pincus, Matthew and Libien, Jenny and others},
	year={2018},
	publisher={Springer}
}

@article{rongala2021non,
	title={A non-spiking neuron model with dynamic leak to avoid instability in recurrent networks},
	author={Rongala, Udaya B and Enander, Jonas MD and Kohler, Matthias and Loeb, Gerald E and J{\"o}rntell, Henrik},
	journal={Frontiers in computational neuroscience},
	volume={15},
	pages={656401},
	year={2021},
	publisher={Frontiers Media SA}
}

@article{stella2015self,
	title={The self-organization of grid cells in 3D},
	author={Stella, Federico and Treves, Alessandro},
	journal={Elife},
	volume={4},
	pages={e05913},
	year={2015},
	publisher={eLife Sciences Publications, Ltd}
}

@inproceedings{NEURIPS2019_6e7d5d25,
	author = {Sorscher, Ben and Mel, Gabriel and Ganguli, Surya and Ocko, Samuel},
	booktitle = {Advances in Neural Information Processing Systems},
	editor = {H. Wallach and H. Larochelle and A. Beygelzimer and F. d\textquotesingle Alch\'{e}-Buc and E. Fox and R. Garnett},
	publisher = {Curran Associates, Inc.},
	title = {A unified theory for the origin of grid cells through the lens of pattern formation},
	volume = {32},
	year = {2019}
}

@article{tani2014self,
	title={Self-organization and compositionality in cognitive brains: A neurorobotics study},
	author={Tani, Jun},
	journal={Proceedings of the IEEE},
	volume={102},
	number={4},
	pages={586--605},
	year={2014},
	publisher={IEEE}
}

@article{pearlmutter1995gradient,
	title={Gradient calculations for dynamic recurrent neural networks: A survey},
	author={Pearlmutter, Barak A},
	journal={IEEE Transactions on Neural networks},
	volume={6},
	number={5},
	pages={1212--1228},
	year={1995},
	publisher={IEEE}
}

@book{tani2016exploring,
	title={Exploring robotic minds: actions, symbols, and consciousness as self-organizing dynamic phenomena},
	author={Tani, Jun},
	year={2016},
	publisher={Oxford University Press}
}

@article{regazzoni2024learning,
	title={Learning the intrinsic dynamics of spatio-temporal processes through latent dynamics networks},
	author={Regazzoni, Francesco and Pagani, Stefano and Salvador, Matteo and Dede’, Luca and Quarteroni, Alfio},
	journal={Nature Communications},
	volume={15},
	number={1},
	pages={1834},
	year={2024},
	publisher={Nature Publishing Group UK London}
}

@article{ctralie2018ripser,
  doi = {10.21105/joss.00925},
  url = {https://doi.org/10.21105/joss.00925},
  year  = {2018},
  month = {Sep},
  publisher = {The Open Journal},
  volume = {3},
  number = {29},
  pages = {925},
  author = {Christopher Tralie and Nathaniel Saul and Rann Bar-On},
  title = {{Ripser.py}: A Lean Persistent Homology Library for Python},
  journal = {The Journal of Open Source Software}
}

@article{Bauer2021Ripser,
    AUTHOR = {Bauer, Ulrich},
     TITLE = {Ripser: efficient computation of {V}ietoris-{R}ips persistence
              barcodes},
   JOURNAL = {J. Appl. Comput. Topol.},
  FJOURNAL = {Journal of Applied and Computational Topology},
    VOLUME = {5},
      YEAR = {2021},
    NUMBER = {3},
     PAGES = {391--423},
      ISSN = {2367-1726},
   MRCLASS = {55N31 (55-04)},
  MRNUMBER = {4298669},
       DOI = {10.1007/s41468-021-00071-5},
       URL = {https://doi.org/10.1007/s41468-021-00071-5},
}

@article{zomorodian_computing_2005,
  title = {Computing {{Persistent Homology}}},
  author = {Zomorodian, Afra and Carlsson, Gunnar},
  date = {2005-02-01},
  journaltitle = {Discrete \& Computational Geometry},
  shortjournal = {Discrete Comput Geom},
  volume = {33},
  number = {2},
  pages = {249--274},
  issn = {1432-0444},
  urldate = {2026-03-26},
  langid = {english}
}

@article{ghrist_barcodes_2007,
  title = {Barcodes: {{The}} Persistent Topology of Data},
  shorttitle = {Barcodes},
  author = {Ghrist, Robert},
  date = {2007-10-26},
  journaltitle = {Bulletin of the American Mathematical Society},
  shortjournal = {Bull. Amer. Math. Soc.},
  volume = {45},
  number = {1},
  pages = {61--75},
  issn = {1088-9485, 0273-0979},
  urldate = {2026-03-26},
  langid = {english}
}

@inproceedings{edelsbrunner_topological_2000,
  title = {Topological Persistence and Simplification},
  booktitle = {Proceedings 41st {{Annual Symposium}} on {{Foundations}} of {{Computer Science}}},
  author = {Edelsbrunner, H. and Letscher, D. and Zomorodian, A.},
  date = {2000-11},
  pages = {454--463},
  issn = {0272-5428},
  urldate = {2026-03-26},
  eventtitle = {41st {{Annual Symposium}} on {{Foundations}} of {{Computer Science}}}
}

@article{obayashi2023stable,
	title={Stable volumes for persistent homology},
	author={Obayashi, Ippei},
	journal={Journal of Applied and Computational Topology},
	volume={7},
	number={4},
	pages={671--706},
	year={2023},
	publisher={Springer}
}

@article{berry2020functional,
	title={Functional summaries of persistence diagrams},
	author={Berry, Eric and Chen, Yen-Chi and Cisewski-Kehe, Jessi and Fasy, Brittany Terese},
	journal={Journal of Applied and Computational Topology},
	volume={4},
	number={2},
	pages={211--262},
	year={2020},
	publisher={Springer}
}

@article{gong2026role,
	title={Role of chloride concentration in modulating seizure transitions in excitatory and inhibitory networks},
	author={Gong, Qianchen and Liu, Yingpeng and Zhang, Yan and Zheng, Muhua and Xu, Kesheng},
	journal={Physical Review E},
	volume={113},
	number={3},
	pages={034401},
	year={2026},
	publisher={APS}
}

%
%

\end{document}